\documentclass[pdflatex,sn-mathphys]{sn-jnl}% Math and Physical Sciences Reference Style
\usepackage{caption}
\usepackage{enumitem}
\usepackage{bm}
\usepackage{diagbox}
\usepackage{bbding}
\jyear{2022}%

\theoremstyle{thmstyleone}%
\theoremstyle{thmstyletwo}%

\theoremstyle{thmstylethree}%

\begin{document}
% \title[Article Title]{GridZero: A Novel Real-Time AC Optimal Power Flow Algorithm for Renewable Power Systems}
% \title[Ariticle Title]{Real-time optimization of multi-period optimal power flow through planning-based reinforcement learning for renewable power systems}
\title[ ]{Real-time scheduling of renewable power systems through planning-based reinforcement learning}


\author[1]{\fnm{Shaohuai} \sur{Liu}\email{liush20@mails.tsinghua.edu.cn}}
\equalcont{These authors contributed equally to this work.}
\author[3]{\fnm{Jinbo} \sur{Liu}\email{liu-jinbo@sgcc.com.cn}}
\equalcont{These authors contributed equally to this work.}
\author[1]{\fnm{Weirui} \sur{Ye}}
\author[2]{\fnm{Nan} \sur{Yang}
% \email{yangnan@epri.sgcc.com.cn}
}

% \email{ywr20@mails.tsinghua.edu.cn}
\author[4]{\fnm{Guanglun} \sur{Zhang}}
\author[4]{\fnm{Haiwang} \sur{Zhong}}
\author[4]{\fnm{Chongqing} \sur{Kang}}
\author[4]{\fnm{Qirong} \sur{Jiang}
% \email{qrjiang@tsinghua.edu.cn}
}
\author[2]{\fnm{Xuri} \sur{Song}}
\author*[2]{\fnm{Fangchun} \sur{Di}
\email{difangchun@epri.sgcc.com.cn}
}
\author*[1]{\fnm{Yang} \sur{Gao}}\email{gaoyangiiis@tsinghua.edu.cn}


\affil[1]{\orgdiv{Institute for Interdisciplinary Information Science}, \orgname{Tsinghua University}, \orgaddress{
% \street{Street}, 
\city{Beijing}, \postcode{100084}, 
% \state{}, 
\country{China}}}

\affil[2]{\orgdiv{China Electric Power Research Institute}, \orgname{State Grid Corporation of China}, \orgaddress{
% \street{Street},
\city{Beijing}, 
\postcode{102299}, 
% \state{State}, 
\country{China}}}

\affil[3]{\orgdiv{National Power Dispatching and Control Center}, \orgname{State Grid Corporation of China}, \orgaddress{
% \street{Street},
\city{Beijing}, 
\postcode{100031}, 
% \state{State}, 
\country{China}}}

\affil[4]{\orgdiv{Department of Electrical Engineering}, \orgname{Tsinghua University}, \orgaddress{
% \street{Street},
\city{Beijing}, 
\postcode{100084}, 
% \state{State}, 
\country{China}}}

% \affil[3]{\orgdiv{Department}, \orgname{Organization}, \orgaddress{\street{Street}, \city{City}, \postcode{610101}, \state{State}, \country{Country}}}


\abstract{
The growing renewable energy sources have posed significant challenges to traditional power scheduling.
It is difficult for operators to obtain accurate day-ahead forecasts of renewable generation, thereby requiring the future scheduling system to make real-time scheduling decisions aligning with ultra-short-term forecasts. Restricted by the computation speed, traditional optimization-based methods can not solve this problem. Recent developments in reinforcement learning (RL) have demonstrated the potential to solve this challenge. However, the existing RL methods are inadequate in terms of constraint complexity, algorithm performance, and environment fidelity. We are the first to propose a systematic solution based on the state-of-the-art reinforcement learning algorithm and the real power grid environment.
The proposed approach enables planning and finer time resolution adjustments of power generators, including unit commitment and economic dispatch, thus increasing the grid's ability to admit more renewable energy. 
The well-trained scheduling agent significantly reduces renewable curtailment and load shedding, which are issues arising from traditional scheduling's reliance on inaccurate day-ahead forecasts.
High-frequency control decisions exploit the existing units' flexibility, reducing the power grid's dependence on hardware transformations and saving investment and operating costs, as demonstrated in experimental results. This research exhibits the potential of reinforcement learning in promoting low-carbon and intelligent power systems and represents a solid step toward sustainable electricity generation.
}


\maketitle

Climate change and carbon neutrality have garnered widespread global attention. The significant amount of carbon emissions during electricity production underscores the importance of achieving low-carbon electricity production as a solution to these pressing challenges.
In recent years, wind and solar energy have emerged as promising sources of sustainable electricity. However, the fluctuation patterns of these sources are highly variable, making it challenging to accurately predict their power generation capacity over the long term.
This presents a major challenge for existing power scheduling systems that rely on reliable long-term forecasts and day-ahead calculation, potentially leading to suboptimal or infeasible solutions, including renewable curtailments and blackouts~\cite{enwiki:1103904998,bialek2020does}.

\begin{figure}[h]
  \centering
  \includegraphics[width=0.9\linewidth]{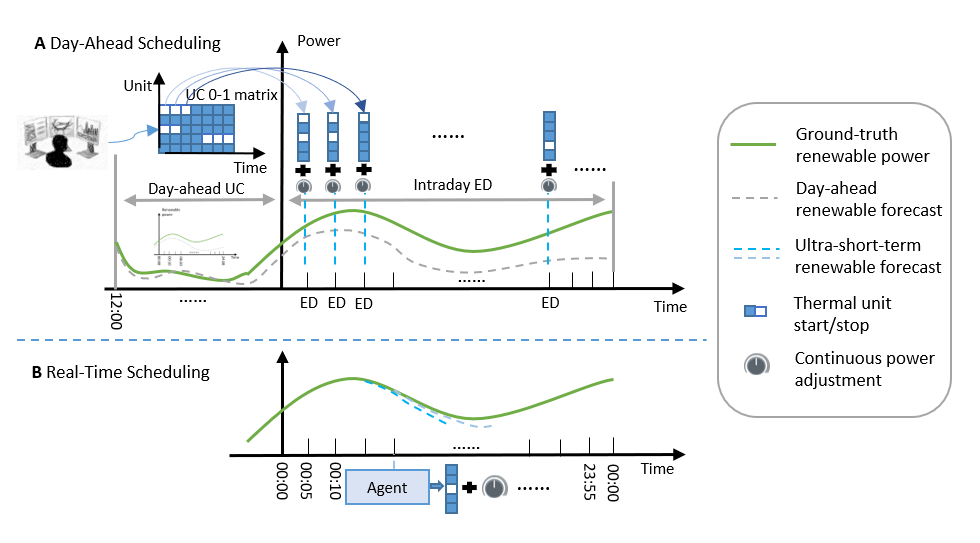}
  \caption{\textbf{DAS vs RTS-GridZero.}
  \textbf{(A) Day-Ahead Scheduling.} DAS performs day-ahead unit commitment to determine the startup/shutdown schedule of thermal generators according to the future 36 hours forecasts, including load and renewable generation. However, the forecasts of renewable generation will become more and more unreliable as the prediction period grows. This introduces errors in the optimization process of the unit commitment, resulting in an unreasonable 0-1 matrix output. Although the coming intraday economic dispatch can partially mitigate the impacts of errors, renewable curtailment and load shedding will occur when the error scale is large.
  \textbf{(B) Real-Time Scheduling.} Conversely, RTS adjusts power generators with a finer time resolution. The control interval is the minute level rather than the day level of the conventional day-ahead scheduling. This is achieved through precise ultra-short-term forecasts and data-driven, computation-efficient reinforcement learning algorithms. Within the narrow period of accurate ultra-short-term forecasts, the scheduling problem is back to a deterministic control problem. The RL agent controls the output power and startup/shutdown of generators simultaneously, corresponding to joint optimization of unit commitment and economic dispatch. Furthermore, our proposed GridZero is capable of look-ahead scheduling for the incoming decision scenarios.
  } 
  \label{fig:das-vs-rts}
\end{figure}

Traditionally, power system operators perform the day-ahead scheduling (DAS) program to calculate power generation schedules~\cite{wood2013power}. 
The DAS program consists of unit commitment (UC) and economic dispatch (ED) as depicted in Fig.\ref{fig:das-vs-rts}(A). 
UC aims at optimizing the combinations of operating power generators to reduce operational costs while maintaining a sufficient power supply. It is an NP-hard problem since it optimizes high-dimensional continuous and integer variables (startup/shutdown of thermal generators) for many time steps~\cite{bendotti2019complexity}.
It is typically solved using the time-consuming mixed integer programming (MIP) method~\cite{bhardwaj2012unit,jabr2012tight}, which could only be calculated a long time in advance. This makes UC heavily rely on accurate forecasts while the day-ahead forecasts of renewable generation are presently unreliable. 
Alternative methods, including reinforcement learning~\cite{de2021applying,de2022reinforcement,ajagekar2022deep}, however, have still been proposed as offline control, nor considered complex network constraints such as transmission line capacity, which is infeasible in actual scheduling. 
On the other hand, ED is usually modeled as a convex optimization problem and can be solved by the interior point and dual methods~\cite{jabr2002primal, olofsson1995linear,sun1984optimal}, but these methods still encounter problems of convergence and computational speed in the face of AC power flow models, in which ED turns into a non-convex problem.
Reinforcement learning methods are also studied to solve ED problems in real-time, while \cite{zhou2020data,zhou2021deep,woo2020real} uses out-of-date algorithms and power grids with a low proportion of renewable energy, which does not accord with future grid development. These RL methods did not consider complex operational constraints such as reactive power limitation and line transmission capacity either.
In general, studying UC or ED alone is inadequate to solve the challenges faced by the power system today.
% How to execute joint optimization at a finer time resolution remains a topic of ongoing research.

The limitations of existing scheduling methods have prompted the adoption of hardware upgrades as alternative solutions, which aimed at increasing flexibility resources, including thermal unit retrofits and energy storage constructions. However, these upgrades come at significant investments. Additionally, if the retrofitted thermal units operate outside of their design specifications, it can lead to elevated carbon emissions and increased costs per megawatt power~\cite{chen2021flexible}. Electrochemical energy storage options, such as batteries, are criticized for their high cost, limited lifespan, and environmental concerns. Physical-based energy storage technologies, such as pumped hydroelectricity and compressed air storage, are also subject to site availability and long construction cycles. Therefore, relying solely on hardware upgrades for enhancing future power scheduling is not a practical and economically viable solution.

We propose to reform the power scheduling system through joint optimization of the above two subproblems with finer time resolutions, referred to as the Real-Time Scheduling (RTS) problem, as demonstrated in Fig.\ref{fig:das-vs-rts}(B). 
This is because recent advancements in deep learning have significantly improved the accuracy of ultra-short-term renewable generation forecasts~\cite{wu2021ultra,tawn2022review}.
The integration of ultra-short-term forecasts enables the transformation of scheduling grids with renewable energy into a deterministic problem within a limited time horizon.
However, traditional optimization algorithms face challenges in generating schedule plans within this short time frame. 
Therefore, there is a need for the development of innovative scheduling algorithms and a shift towards a computationally efficient framework, which is capable of real-time optimization of unit commitment and economic dispatch simultaneously.

In contrast to optimizing UC and ED separately, we take a drastically different technical approach to real-time joint optimization and take all system operational constraints into consideration.
We propose to take advantage of reinforcement learning (RL) to solve the problem in real-time by shifting the intensive computational burden of traditional optimization algorithms to an offline training process. 
RL has shown its capabilities to solve complex control tasks while remaining computationally efficient\cite{mnih2015human,lillicrap2015continuous,haarnoja2018soft,schulman2017proximal,silver2016mastering}. Some aspects of power systems, such as full-fledged simulation and wide-spread smart measurements, are suitable for RL's applications.
However, we find that the most widely used RL algorithms, such as Deep Deterministic Policy Gradient (DDPG)\cite{lillicrap2015continuous}, Soft Actor-Critic (SAC)\cite{haarnoja2018soft}, and Proximal Policy Optimization (PPO)\cite{schulman2017proximal} are unable to achieve satisfactory results. As a result, we developed a look-ahead scheduling RL method called GridZero, built upon the state-of-the-art AlphaGo series' work, which excels in complex board games~\cite{silver2016mastering,li2018alphago,silver2017mastering,schrittwieser2020mastering}, robotic control~\cite{hubert2021learning}, scientific discovery~\cite{fawzi2022discovering}, and video compression~\cite{mandhane2022muzero}.
The choice of this search plus deep learning framework for solving the RTS task is based on the common feature between power scheduling and Go games, which both require the look-ahead search for future possible situations. The search method in the AlphaGo series has been demonstrated to produce robust and conservative policies, which are critical for power dispatching, as security is the first criterion of the power grid.

The simulation results of the real power grid demonstrate the efficacy of our proposed method in addressing the major challenges hindering the widespread adoption of renewable energy. Specifically, our method enables fast scheduling, reduces renewable generation curtailment, eliminates load shedding, and minimizes the need for expensive hardware upgrades. In the test scenario with a high share of renewable generation, our method reduced renewable generation curtailment by 79\% and eliminated load shedding - issues that are commonly encountered in DAS due to unreliable forecasts. By executing control at finer time resolution, our method conserves the cost of hardware upgrades, which would have required an upgraded capacity of 20\% of the total installed capacity, equating to a minimum investment of 60 million dollars. This amount of savings is sufficient to build 14 100-MW coal-fired power plants. In conclusion, our RL-based method guarantees stable and efficient real-time scheduling in power systems with high renewable energy penetration.

\begin{figure}[h]
  \centering
  \includegraphics[width=0.9\linewidth]{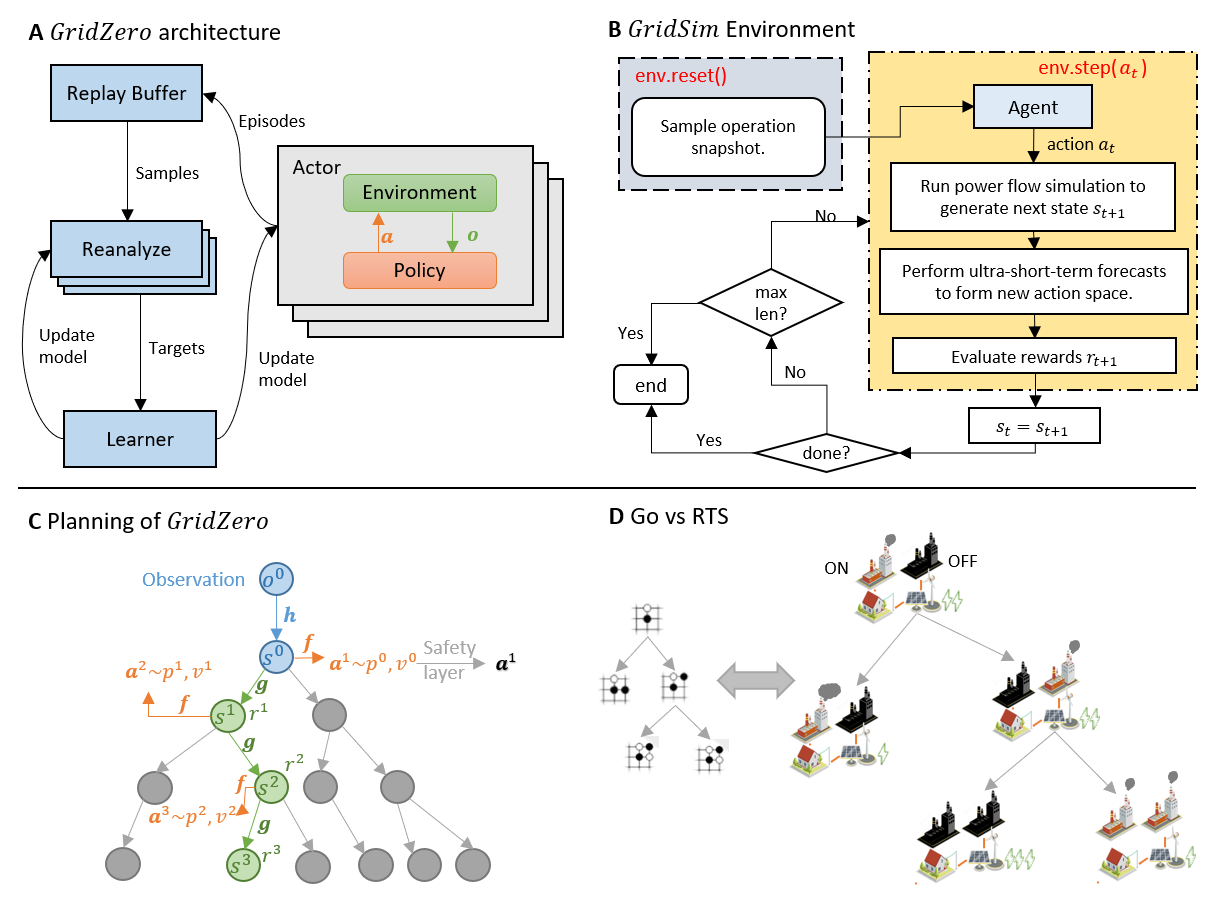}
  \caption{\textbf{Illustration of GridZero.}
  \textbf{(A) Architecture.}  
  Actors interact with GridSim to collect state transitions which are then stored in a replay buffer. The replay buffer feeds data to reanalyze to make targets for the learner.
  \textbf{(B) Flow chart of power scheduling RL environment.} 
  GridSim simulates the next state through power flow computation, performs forecasts for load and renewable generation capacity, and calculates the reward.
  % includes components such as renewable and load prediction, power flow simulation, and reward calculation.
  \textbf{(C) Look-ahead scheduling.} GridZero grows a search tree, with the observation $o$ encoded as root hidden state $s$ by the representation network $h$. During each simulation, the algorithm descends to a leaf node along the most promising path and adds a new leaf. The dynamics network $g$ estimates the hidden state $s$ and reward $r$ of each new node, while the prediction network $f$ provides the value $v$ and policy $p$. The action is selected from the root visit distribution and may be adjusted to ensure safety.
  \textbf{(D) Go vs RTS}. AlphaGo's success in the game of Go benefits from the look-ahead search for both sides' actions. The search mechanism is beneficial for the agent to search for a future step that is far beyond what a human can think about and to make the most rational decision. This idea is also suited for power scheduling since the operators need to simulate multi-step future scenarios and make the most reasonable scheduling plan. 
  } 
  \label{fig:nature_fig1}
\end{figure}

\subsection*{Real-time scheduling as an RL problem}
Optimization-based methods are time-consuming, resulting in a reliance that traditional scheduling methods can only be performed through prior computation.
This can lead to scheduling results susceptible to errors in day-ahead forecasts of renewable generation. 

Real-time scheduling (RTS) addresses this challenge by utilizing precise ultra-short-term forecasts to solve a nearly deterministic problem. 
Reinforcement learning (RL) is a compatible solution for RTS, as it shifts the time-intensive computation to the offline training process, exhibiting the ability to resolve other complex control tasks~\cite{silver2016mastering,lillicrap2015continuous,schulman2017proximal,degrave2022magnetic,fawzi2022discovering,mandhane2022muzero}.
In this study, the RTS problem is reformulated as a sequential Markov decision process (MDP). 
The proposed scheduling MDP incorporates ultra-short-term forecasts as components of the observation. It considers all operational constraints and optimization objectives of the power system operation in the reward function, guiding the agent to improve the scheduling objectives while adhering to operational constraints. 
This MDP serves as a foundation for deploying computationally efficient RL algorithms.

The proposed RL-based RTS framework is illustrated in Fig.\ref{fig:nature_fig1}. Our approach encompasses two major phases. Initially, the RTS issue is represented as a Markov Decision Process (MDP) and an RL environment, referred to as GridSim, is constructed. Subsequently, a look-ahead scheduling RL agent is introduced that interacts with GridSim to learn near-optimal control policies through a systematic search and training procedure.

In the initial phase, 
the MDP and Gridsim environment are modeled as follows:
\begin{enumerate}[label=(\arabic*)]
    \item Observations encode the grid's operational states, including generator power outputs, load consumption power, line transmission currents, bus voltages, etc. Observations also include the next-step prediction of renewable maximum power and load consumption.
    \item Actions are designed as concatenations of generator active power outputs' continuous adjustments and discrete unit startup/shutdowns, 
    which aims to optimize the economic dispatch and unit commitment simultaneously. Furthermore, renewable generators do not always output full power. Their power setpoint range is $[0, \overline{\text{p}^t}]$ where $\overline{\text{p}^t}$ is the maximum power of renewable generators at step $t$. Renewable curtailment is the difference between the maximum power and the actual power of renewable generators.
    \item State transitions represent the conversion to the next power system steady state, and they are modeled in AC power flow models and simulated through a professional power flow analysis program, which solves the power flow equations via the Newton-Raphson method. 
    \item The reward is a scalar function that measures the grid's current state under the criteria of specific constraints and objectives. It also penalizes actions that lead to undesirable states that violate operational constraints.
\end{enumerate}

In the second phase, an RL algorithm called GridZero is designed to make better use of the look-ahead search's benefits in solving the RTS problem. GridZero interacts with GridSim and accumulates state transition data to develop policies.
The architecture of tree search combined with deep neural networks delivers more stable policy improvements compared to commonly utilized model-free RL methods such as\cite{lillicrap2015continuous,haarnoja2018soft}. This enables GridZero to attain satisfactory scheduling performances in the RTS task. 
The capability comparison between GridZero and conventional methods is presented in Table.\ref{tab:capability}. GridZero represents a groundbreaking RL-based method that not only satisfies the real-time requirement and utilizes ultra-short-term forecasting but also possesses the ability to plan for future scenarios.

\begin{table}[h]
\centering
\caption{Comparisons of DAS, Model-free RL, and GridZero. GridZero achieves look-ahead scheduling, utilization of ultra-short-term forecasts, and real-time scheduling simultaneously.}
% \begin{tiny}
\begin{tabular}{cccc}
    \toprule
      & DAS & Model-free RL & GridZero \\
    \midrule
    Look-ahead scheduling & \CheckmarkBold & \XSolidBold & \CheckmarkBold \\
    Ultra-short-term forecasts & \XSolidBold & \CheckmarkBold & \CheckmarkBold \\
    Real-time decision-making & \XSolidBold & \CheckmarkBold & \CheckmarkBold \\
    \bottomrule
  \end{tabular}
% \end{tiny}
  \label{tab:capability}
\end{table}

As depicted in Fig.\ref{fig:nature_fig1}(C), a systematic search process characterized by a growing search tree, is employed to find a carefully selected action candidate. This process evaluates the grid scheduling actions by simulating how the grid will change after the proposed actions, facilitated by the dynamic function. Through a hybrid sampling process, the policy function proposes action candidates to narrow the search process to promising future scenarios.
The value function then assesses the preference of states encountered during the search process, by estimating the discounted future reward. The UCB score, a combination of value prediction and node visited number, balances exploration and exploitation, thus guiding the selection process in the tree search. Intuitively, the search process could avoid the case when the current operation is valid but leads to future grid states that are hard to handle. It is reminiscent of how an experienced power grid manager operates the grid, as they intuitively know some good candidate controlling options according to their past experience, and then they validate the operation by further simulations.

Meanwhile, we propose a safety layer to prevent the RL agent from causing destructive consequences due to unrestricted exploration in this risk-sensitive task.
Purely random explorations can lead to severe consequences, as randomly setting the generators' power outputs cannot guarantee the basic load-generation balance.  
This is particularly concerning in power systems because any violations of this constraint can result in blackouts or even system failures\cite{enwiki:1103904998,bialek2020does}.
To simulate real scenarios, GridSim implemented such violations of power balancing as immediate episode terminations, which makes it difficult for agents to continuously interact with GridSim while exploring without limit.
Additionally, the temporal varieties of renewable generation make the action space change over time. 
To satisfy this time-varying power balance constraint, the safety layer provides a feasible solution space, which enables the RL agent to explore safely in it. 
Such a safety layer is common in many other similar tasks with high-security requirements\cite{shao2022safety,chitta2022transfuser}.

\subsection*{GridSim simulates real scheduling scenarios} 
The experiments are conducted within the GridSim environment, which is modeled after a real power grid of a province of China.
The proportion of renewable generators in this test grid constitutes one-third of the total number of generators, and in certain scenarios, the maximum power generated by renewable units can surpass 60\% of the load. These align with the envisioned structure of future power systems with high penetration of renewable energy.

The core of GridSim is a professional power flow analysis program, solving power flow equations typically via the Newton-Raphson method\cite{wood2013power}. This professional program is widely employed in practical operations of the State Grid Corporation of China, enabling GridSim to boast sufficient physical fidelity to describe the complex state transitions of power systems. The provincial power grid, named SG-126, has 126 buses, 194 lines, 91 loads, and 54 generators, 17 of which are renewable.

GridSim encompasses a whole year's operational snapshots, which enables RL agents to interact with the environment at any time of the year, thus facilitating the learning of robust and generalized scheduling policies.
These operational snapshots, which comprise load, renewable generation, and generators' power set-points, are derived from actual measurements. They were logged during a full year with data points spaced at 5-minute intervals, resulting in 105,120 operational snapshots. Additionally, the training and test datasets are separated, comprising measured snapshots of 2 consecutive years. This ensures that the agent learns generalized scheduling strategies and captures the essential dynamic characteristics of the grid operation instead of overfitting the training dataset.

\subsection*{General scheduling capability demonstration}

\begin{figure}[h]
  \centering
  \includegraphics[width=1.0\linewidth]{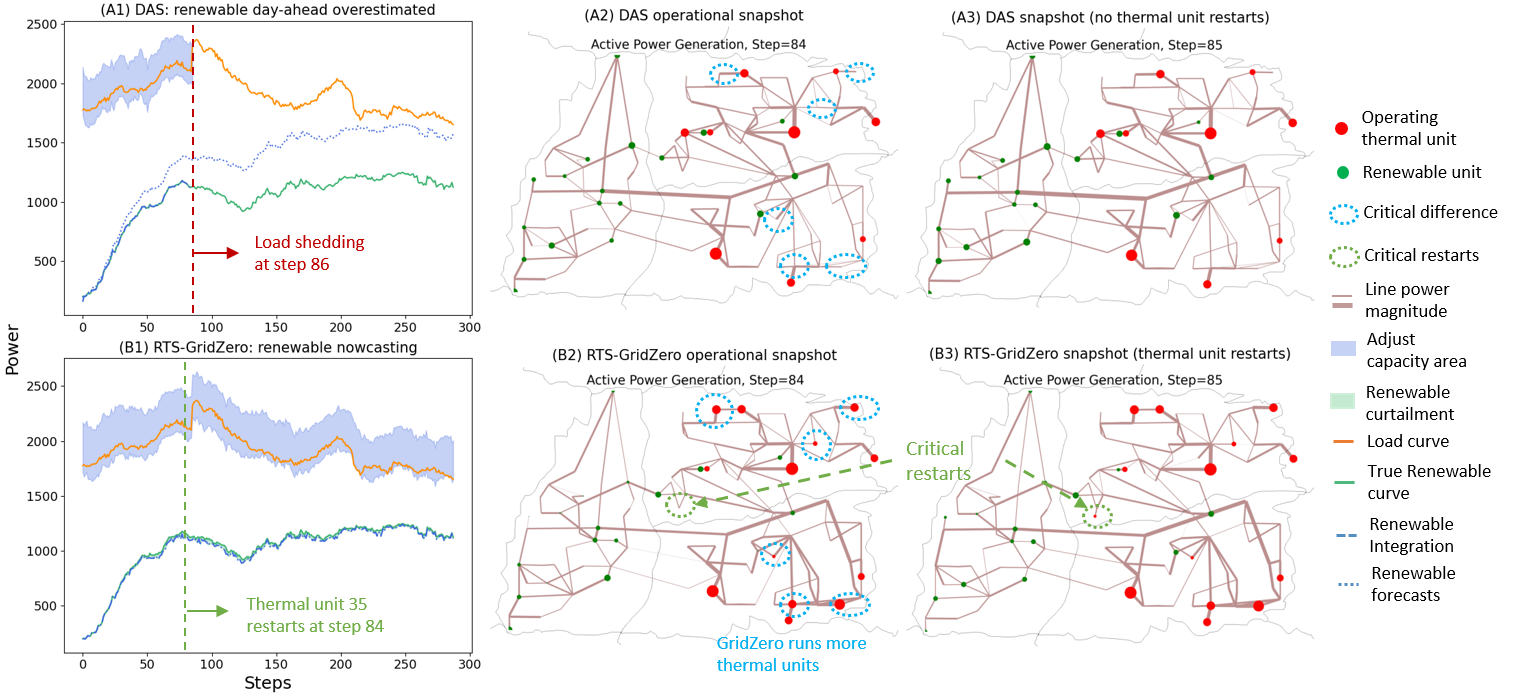}
  \caption{\textbf{RTS-GridZero reduces load shedding caused by renewable overestimation in the DAS framework.} Figures (A*) series indicate the performance of DAS under overestimated renewable forecasts, and figures (B*) series represent the performance of RTS-GridZero. 
  The red nodes indicate operating thermal units, and the green nodes represent operating renewable units. 
  The node sizes represent the magnitudes of active power generation. The line thicknesses represent the magnitudes of transmission power. The visualizations of grid operational snapshots are conducted in PYPSA~\cite{brown2017pypsa}. 
  As shown in (A1), DAS encounters a power mismatch when faced with a load surge at step 86. This is because that DAS relies on inaccurate day-ahead forecasts of renewable generation, leading to insufficient operating thermal units as marked by blue circles in (A2) and (B2). By using ultra-short-term forecasts and the ability to make look-ahead scheduling, GridZero first maintains a reasonable number of operating thermal units, which ensures sufficient ramping power if faced with load surges. Second, GridZero restarts thermal unit 35 at step 84, as marked by green circles in (B2) and (B3). These ensure that GridZero has adequate ramping power to handle the coming load surge at step 86.
  } 
  \label{fig:day_ahead_uc}
\end{figure}

\begin{figure}[h]
  \centering
  \includegraphics[width=1.0\linewidth]{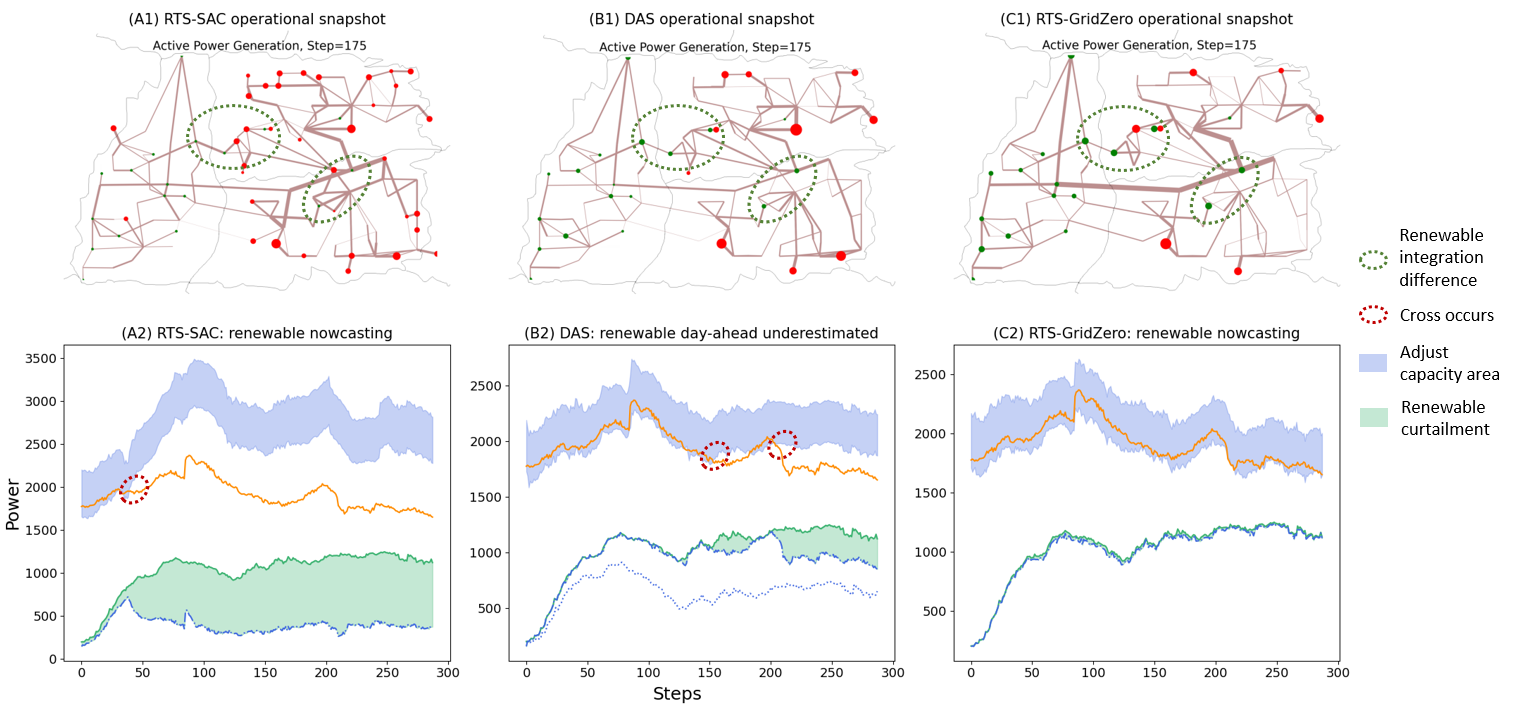}
  \caption{\textbf{RTS-GridZero outperforms SAC and DAS in reducing renewable curtailment.}
  SAC runs excessive thermal units leading to abundant renewable curtailment. DAS is also affected by underestimated renewable forecasts and maintains an excessive number of operating thermal units. These caused renewable curtailments since the lower bounds of the adjust capacity areas surpass the load curve, as marked by red circles in (A2) and (B2). With the powerful policy improvement and look-ahead scheduling brought by MCTS, GridZero learns efficient scheduling policies and integrates more renewable generation than SAC and DAS, as marked in green circles in (A1), (B1), and (C1). This significantly reduces renewable curtailment as shown in (C2). 
  % The red nodes indicate operating thermal units, and the green nodes represent operating renewable units. The node size represents the value of active power generation. The line thickness represents the amplitude of transmission power. 
  } 
  \label{fig:renewable_curtailment}
\end{figure}

To evaluate the performance of GridZero, DAS, and SAC, a challenging test scenario is devised with a renewable generation proportion reaching up to 60\%. We choose SAC as the baseline of the model-free method because SAC performs better than DDPG and PPO in this scheduling problem.
The test scenario involves a whole-day scheduling task with 288 decision steps taken at 5-minute intervals. 
DAS is developed through the resolution of day-ahead unit commitment using the optimization software GUROBI and intraday economic dispatch using PYPOWER~\cite{lincoln2019pypower}. 
To demonstrate the advantages of our approach over traditional DAS, we first compare GridZero and DAS in reducing load shedding, as shown in Fig.\ref{fig:day_ahead_uc}. Second, we compare the performance of SAC, DAS, and GridZero in reducing the renewable curtailment, as demonstrated in Fig.\ref{fig:renewable_curtailment}.
SAC is also equipped with the proposed power balancing safety layer for effective exploration and is reproduced using the distributed APEX architecture which allows efficient data collection through multiple processes~\cite{horgan2018distributed}.

The DAS approach is prone to generating infeasible or suboptimal solutions if receiving inaccurate day-ahead forecasts, particularly in the UC problem.
If the renewable generation capacity is overestimated, as demonstrated in Fig.\ref{fig:day_ahead_uc}(A1), it can mislead DAS and result in insufficient operating thermal generators, as marked by blue circles in Fig.\ref{fig:day_ahead_uc}(A2). This is because that DAS's results are typically greedy, which makes it sensitive to prediction errors. The insufficient operating units caused a shortage in ramping power and a mismatch between generation and load power at step 86, as shown in Fig\ref{fig:day_ahead_uc}(A1). In real-world power grid operations, such power mismatches affect the frequency stability, leading to load shedding or even more serious consequences, such as grid collapse\cite{enwiki:1103904998}. However, GridZero solves this problem by look-ahead scheduling, it doesn't optimize the operational cost greedily but chooses the scheduling decision with a high expected return in the future, in which robustness has a greater impact. GridZero maintains a reasonable number of operating thermal units, as marked by blue circles in Fig.\ref{fig:day_ahead_uc}(B2). It anticipates the coming load surge in the look-ahead search process and restarts the thermal generator 35 at step 84 in advance, as marked by green circles in \ref{fig:day_ahead_uc}(B2) and (B3).

On the other hand, the underestimated renewable generation capacity also misleads DAS to run an excessive number of operating thermal generators, as shown in Fig.\ref{fig:renewable_curtailment}(B1). This causes the lower bound of the adjust capacity area to surpass the load consumption, as marked by red circles in Fig.\ref{fig:renewable_curtailment}(B2), resulting in the curtailment of renewable integration. This, in turn, increases carbon emissions and the costs of grid operations. 
The adjustment capacity area is based on the assumption of full integration of renewable generation.
% Essentially, when the load curve crosses the upper and lower bounds of the adjustment capacity area, load shedding and renewable curtailment will occur respectively. 

Model-free RL methods, such as SAC, are commonly criticized for potentially inaccurate value estimations. This can cause instability in policy improvements and the scheduling agent is not able to integrate renewable generation reasonably, referred to as abundant renewable curtailment. As demonstrated in Fig.\ref{fig:renewable_curtailment}(A1) and (A2), SAC employs an excessive number of thermal generators starting from step 40, causing the adjustment capacity area to significantly surpass the load curve, as marked by the red circle in \ref{fig:renewable_curtailment}(A2). This in turn leads to a significant amount of renewable generation being curtailed.

However, with the utilization of ultra-short-term predictions and a well-designed look-ahead scheduling strategy, GridZero is able to make prompt and rational decisions regarding the power outputs of all generators and startup/shutdowns of thermal units, thereby ensuring that the adjustment capacity area aligns with the load consumption curve. This observably reduces renewable curtailment, as demonstrated in Fig.\ref{fig:renewable_curtailment}(C1) and (C2). 

\subsection*{Quantative analysis of GridZero and other methods}

\begin{table}[h]
\centering
\caption{Test statistics of GridZero, SAC, and traditional DAS (4 runs with 20 seeds). The RL-based RTS methods achieve much faster computation speed than the DAS precalculation framework by shifting the computation burden from the optimization process to an offline training phase. GridZero also much outperforms SAC in general performance and better observes operational constraints. $T_\text{episode}$ is 288-step decision-making time of a whole day. $\vert V_\text{bus}\vert$ is the voltage magnitude of the bus. $Q$ is the reactive power output of the generator.}
% \begin{tiny}
\begin{tabular}{cccc}
    \toprule
      & SAC & DAS & \textbf{GridZero} \\
    \midrule
    $T_{\text{episode}}$(s) & $21.2\pm{1.8}$ & $8557.3\pm{304.1}$ & $43.2\pm{2.1}$ \\
    Cumulative rewards & $256.8\pm{51.3}$ & 
    % $649.4\pm{0.12}$ 
    -
    & $\textbf{628.3}\pm{35.2}$ \\
    $\vert V_\text{bus}\vert$ violation(\%) & $0.3\pm{0.2}$ & $0.8\pm{0.1}$ & $0.5\pm{0.1}$ \\
    $Q$ violation(\%) & $15\pm{7}$ & $10\pm{1}$ & $\textbf{6.5}\pm{2}$ \\
    $P_{\text{balanced}}$ violation(\%) & $0.2\pm{0.3}$ & $0\pm{0.1}$ & $\textbf{0.05}\pm{0.05}$ \\
    Line soft overflow(\%) & $6\pm{4}$ & $8.8\pm{0.3}$ & $\textbf{3}\pm{3}$ \\
    Line hard overflow(\%) & $0.5\pm{0.1}$ & $0.1\pm{0.1}$ & $\textbf{0}\pm{0.01}$ \\
    Operating cost & $51135\pm{4000}$ & $41237\pm{200}$ & $\textbf{39785}\pm{2500}$ \\
    Renewable consumption(\%) & $55\pm{5}$ & $89\pm{0.5}$ & $\textbf{95}\pm{2}$\\
    \bottomrule
  \end{tabular}
% \end{tiny}
  \label{tab:statitics}
\end{table}

\begin{figure}[h]
  \centering
  \includegraphics[width=1.0\linewidth]{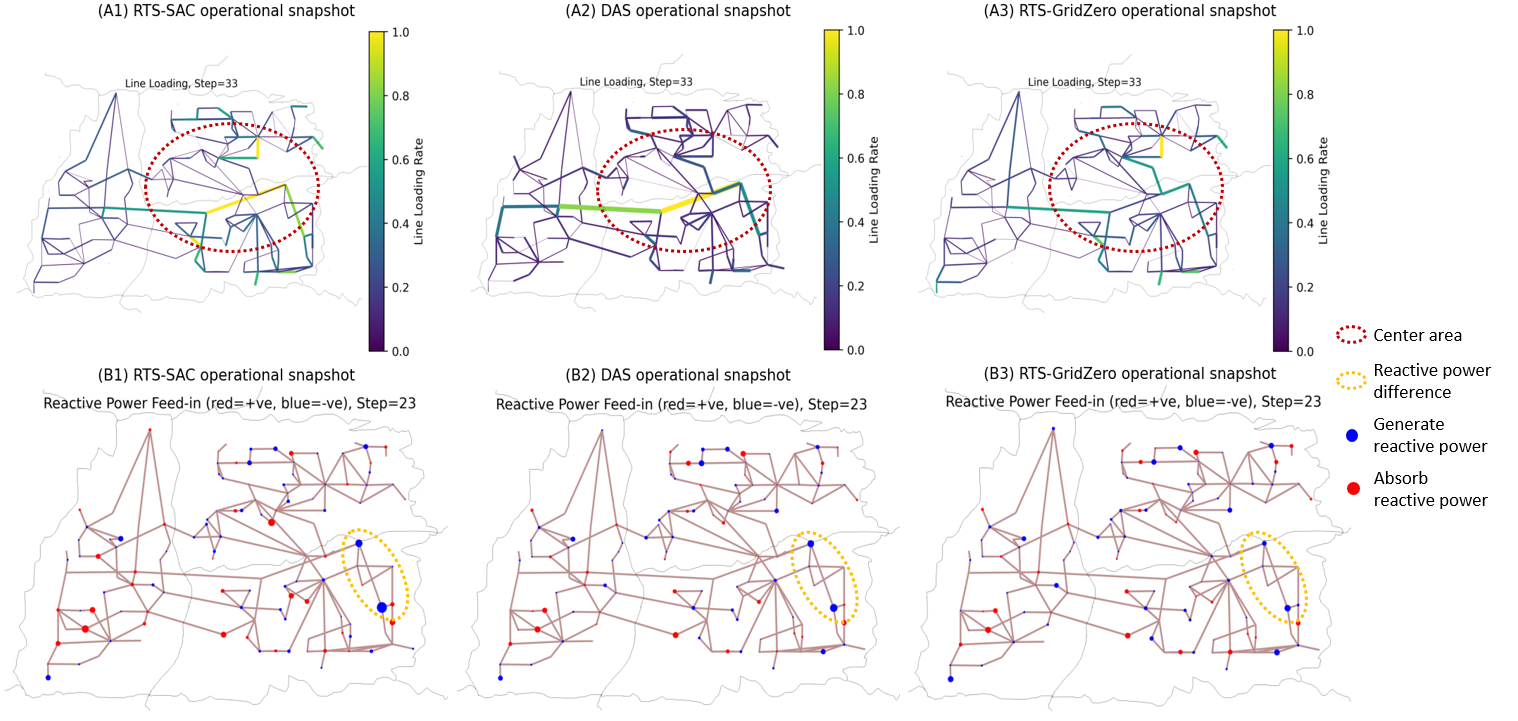}
  \caption{\textbf{Visualizations of line loading and reactive power dispatching.}
  Figures (A*) series indicate the operational snapshots of line loading, and figures (B*) series represent the operational snapshots of reactive power. For line loading, a reasonable strategy is to balance the load rate of each line and to avoid the situation that line loading is heavy. As marked by red circles in (A1), (A2), and (A3), GridZero achieves more average line loading than DAS and SAC in the center area. For reactive power, a reasonable strategy is to avoid reactive power converging on a small number of buses, which might lead to reactive power out-of-limit. GridZero also alleviates the reactive power convergence as marked by yellow circles in (B1), (B2), and (B3). The red nodes absorb reactive power and the blue nodes generate reactive power. 
  } 
  \label{fig:overflow_reactive}
\end{figure}

In addition to superior performance in representative scenarios, GridZero outperforms both DAS and SAC in various scheduling quality metrics.
As shown in Table.\ref{tab:statitics}, our method achieves 188.9 times faster than the conventional DAS method, corresponding to 0.15 seconds per decision step, which ensures the ability of real-time scheduling. DAS solves a mixed integer programming problem with 10080 discrete variables and 15552 continuous variables in a whole-day scheduling task, which is highly time-intensive. GridZero also outperforms SAC and DAS in terms of constraints and operating costs. 
The score of DAS is not shown because DAS is not capable of making single-step decisions like RL agents in GridSim. 
Our approach leverages ultra-short-term forecasts to make instant adjustments to generators, providing greater flexibility than traditional methods that rely solely on day-ahead plans. This allows the grid to make better use of renewable energy sources and minimize power generation costs.

% As demonstrated in Table.\ref{tab:statitics}, our method is superior to both DAS and SAC in terms of line overflow rate and reactive power out-of-limit.
The line loading is visualized in Fig.\ref{fig:overflow_reactive}(A*). GridZero maintains more average line loading, which could help reduce transmission overflows and line outages. In Fig.\ref{fig:overflow_reactive}(B*), GridZero also achieves more balanced reactive power dispatch and avoids reactive power convergence in Fig.\ref{fig:overflow_reactive}(B1). This can help reduce the violations of generator reactive power limits.
% In addition to the power balancing constraint addressed by the proposed safety layer, the remaining constraints are learned through interactions with the environment.
% Although these constraints may be violated during exploration processes, they can also occur in actual grid operations.
% The rewards are set as negative penalties for violations, and small positive bonuses for satisfying obedience to the constraints. 
% This learning approach allows the agent to understand the power system's operation rules through the penalties for violating the constraints, and to balance complex constraints with optimization objectives. This is a more effective method than accelerating traditional optimization algorithms, which often omit constraints in the pursuit of computing speed.

\subsection*{GridZero reduces the cost of flexibility retrofits and energy storage}
Real grid simulation results show that GridZero is able to achieve efficient real-time scheduling without the need for additional flexible resources, resulting in lower levels of renewable curtailment and load shedding. On the other hand, maintaining the DAS framework without changes necessitates significant hardware upgrades to enhance the grid's flexible resources. Such hardware improvements often require substantial investments.
 
More specifically, the hardware upgrades required for the DAS framework to reduce renewable curtailment and load shedding can be divided into two main directions, namely thermal units' flexibility retrofit and energy storage construction.
The objective of thermal units' flexibility retrofit is to lower the minimum operating power of thermal generators. 
In this way, even if the renewable energy power generation is underestimated, resulting in excessive thermal power generators operating, the output power of thermal generators can be reduced sufficiently to avoid renewable energy curtailment.
Energy storage serves a similar purpose, reducing the mismatch between renewable energy and load by storing excess renewable power during off-peak periods and releasing the stored electricity when the power supply is tight.
Both these directions involve significant investments. 

% \begin{table}[h]
% \centering
% \caption{Cost analysis of 
% % Thermal Flexibility Retrofits, Electro-Chemical Storage, and Pumped Hydroelectricity Storage, 
% DAS with hardware upgrades and RTS-GridZero under different levels of renewable energy penetration,
% in billions of dollars. 
% In the case of renewable energy penetration of 30\%, the capacity of flexibility retrofit or energy storage should reach 10\% of the total installed capacity. While in the case of 60\% penetration, the capacity of flexibility retrofit and energy storage should reach more than 20\%.
% Thermal unit Flexibility Retrofit (TFR) is currently the most cost-effective method, compared with Electro-Chemical Storage (ECS) and Pumped Hydroelectricity Storage (PHS). GridZero simply improves on the scheduling algorithm and does not require additional hardware upgrades, which can save grid operators significant amounts of investments.}
% \begin{tiny}
% \begin{tabular}{ccccc}
%     \toprule
%       Cost (Million \$)
%       & DAS-TFR & DAS-ECS & DAS-PHS & \textbf{RTS-GridZero} \\
%     \midrule
%     RES penetration 30\% & \multirow{2}*{29.9} & \multirow{2}*{257.2} & \multirow{2}*{155.1} & \multirow{2}*{\textbf{0.0}} \\
%     (retrofit capacity 10\%, 201.1MW) & ~ & ~ & ~ & ~\\
%     \midrule
%     RES penetration 60\% & \multirow{2}*{60.1} & \multirow{2}*{514.7} & \multirow{2}*{309.7} & \multirow{2}*{\textbf{0.0}} \\
%     (retrofit capacity 20\%, 402.2MW) & ~ & ~ & ~ & ~ \\
%     \bottomrule
%   \end{tabular}
% \end{tiny}
%   \label{tab:cost}
% \end{table}

\begin{figure}[h]
  \centering
  \includegraphics[width=1.0\linewidth]{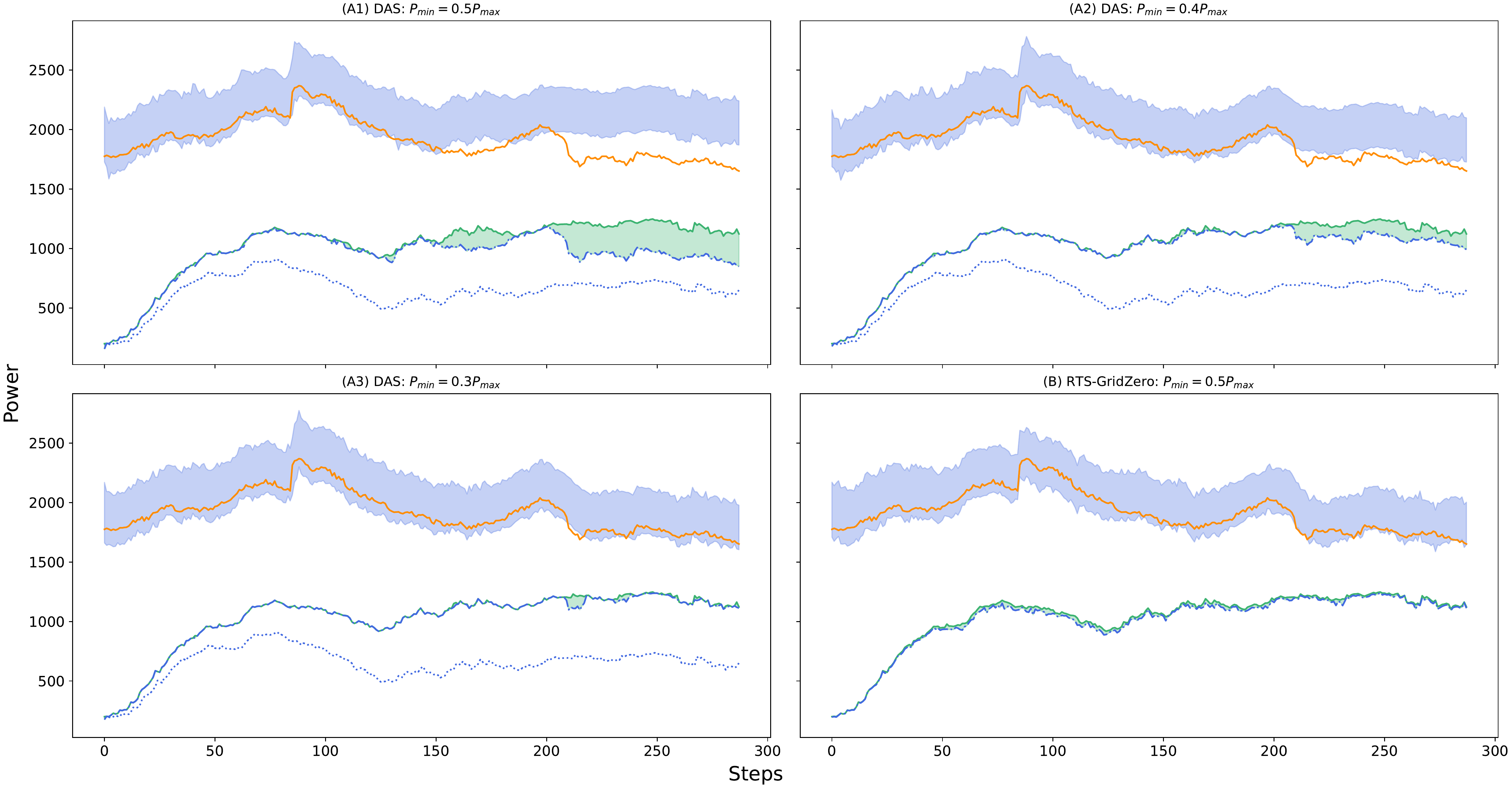}
  \caption{\textbf{Impacts of different flexibility retrofit levels on the scheduling performance.} 
  (A1),(A2) and (A3) shows the flexibility retrofit levels of 
  0\%, 10\% and 20\%, which are corresponding to $\underline{\text{p}}=0.5\,\overline{\text{p}}, 0.4\,\overline{\text{p}}, 0.3\,\overline{\text{p}}$. $\overline{\text{p}}$ and $\underline{\text{p}}$ are the maximum power and minimum power of thermal generators.
  Their results are all based on the DAS. (B) shows the performance of GridZero with $\underline{\text{p}}=0.5\,\overline{\text{p}}$. $\underline{\text{p}}=0.5\,\overline{\text{p}}$ is the most common design of actual thermal generators. $\underline{\text{p}}=0.4\,\overline{\text{p}}$ and $\underline{\text{p}}=0.3\,\overline{\text{p}}$ correspond to 10\% and 20\% flexibility retrofit levels respectively. As we can see, using the DAS approach requires a 20\% flexibility retrofit of conventional thermal units to significantly reduce renewable curtailment. However, the GridZero-based RTS approach achieves the same result without requiring hardware investments to the existing grid through adjusting generators in real-time according to ultra-short-term forecasts.
  } 
  \label{fig:flexibility_retrofit}
\end{figure}

\begin{figure}[h]
    \centering
    \includegraphics[width=1.0\linewidth]{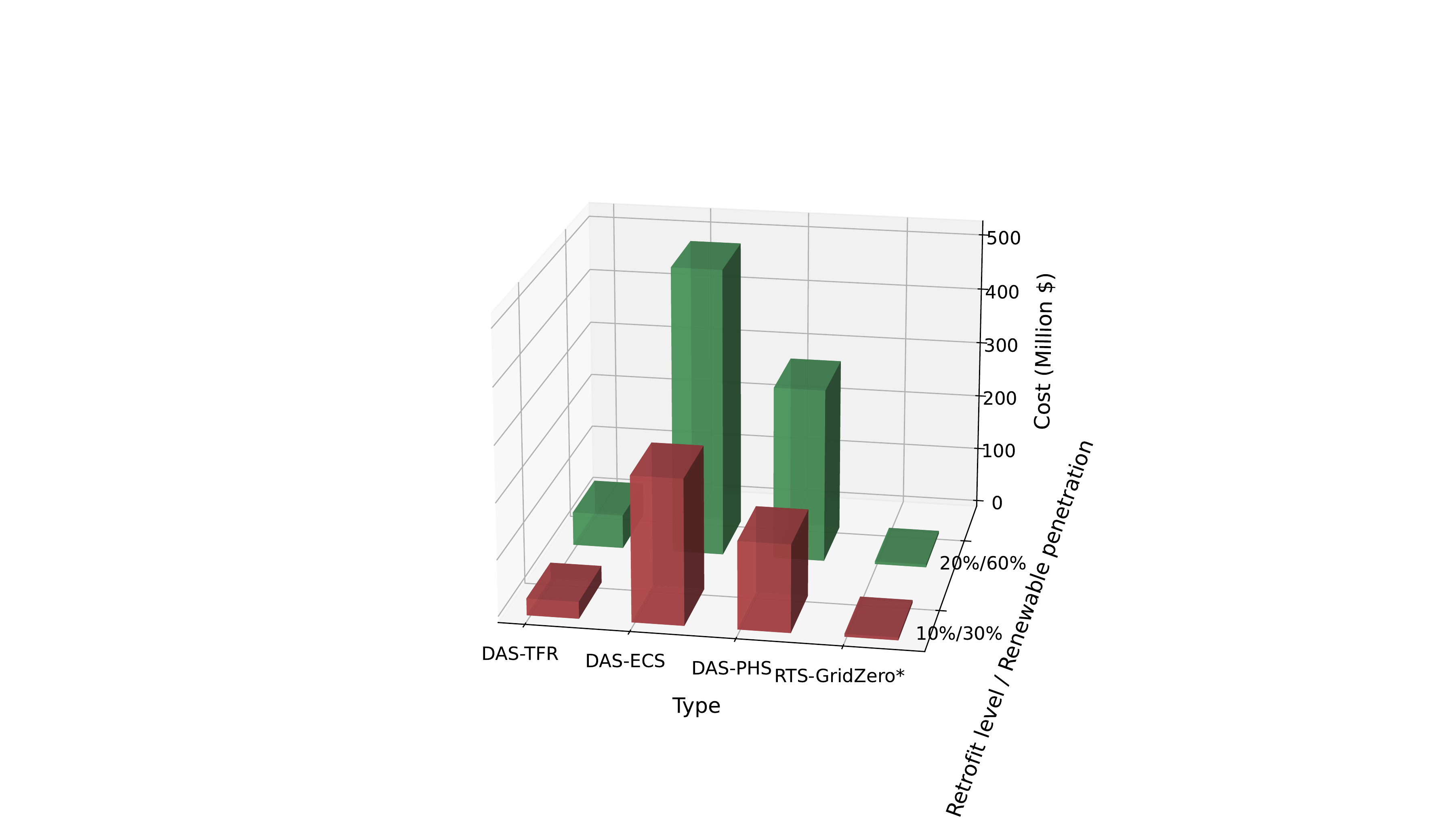}
    \caption{Cost analysis of DAS with hardware upgrades and RTS-GridZero under different levels of renewable energy penetration, in billions of dollars. 
    In the case of renewable energy penetration of 30\%, the capacity of flexibility retrofit or energy storage should reach 10\% of the total installed capacity. While in the case of 60\% penetration, the capacity of flexibility retrofit and energy storage should reach more than 20\%.
    Thermal unit Flexibility Retrofit (TFR) is currently the most cost-effective method, compared with Electro-Chemical Storage (ECS) and Pumped Hydroelectricity Storage (PHS). GridZero simply improves on the scheduling algorithm and does not require additional hardware upgrades, which can save grid operators significant amounts of investments.}
    \label{fig:cost}
\end{figure}

As illustrated in Fig.\ref{fig:flexibility_retrofit}, in order to mitigate renewable curtailment, a retrofit capacity equivalent to 20\% of the total installed capacity is required, which corresponds to 402.2 megawatts of retrofit capacity.
An analysis of China's Jiangxi province indicates that the cost of thermal unit flexibility retrofit is 0.15 million dollars per megawatts~\cite{chen2021flexible}.
Therefore, the total cost of this approach can reach 60.1 million dollars in investment. Additionally, this approach results in the uneconomical operation of thermal units, leading to higher generation costs and increased carbon emissions. According to ~\cite{chen2021flexible}, thermal generators operating at 30\% rated power have 70\% higher cost and 85\% more carbon emission per unit of power generation, compared to those operating at 50\% rated power.

Energy storage is even more expensive than thermal units' flexibility retrofits. The lithium-ion battery is the most cost-effective electrochemical storage choice, but its cost per megawatts is 1.28 million dollars, which is much higher than thermal generator flexibility retrofits~\cite{yan2022lcos}. Although hydro-pumped storage is cheaper than batteries, costing 0.77 million dollars per megawatts~\cite{sospiro2021cost}, 
this physically based long-timescale storage method is still faced with challenges, such as the difficulty of site selection and long construction cycles. As indicated in Fig.\ref{fig:cost}, the cost of constructing energy storage is much higher than the cost of thermal units' flexibility retrofit, despite the same retrofitted level.

In light of these considerations, we investigate the efficacy of our approach in reducing the grid's dependence on energy storage and flexibility retrofits. Fig.\ref{fig:flexibility_retrofit} illustrates that GridZero can achieve a performance level comparable to that of the DAS operating with 20\% retrofitted capacity.
Notably, as Fig.\ref{fig:cost} summarizes, GridZero does not require thermal units' flexibility retrofits or energy storage construction, even if the proportion of renewable energy power generation increases to 60\%. Our methodology relies solely on ultra-short-term forecasts and fast adjustments to enhance dispatch flexibility, which can translate into substantial cost savings for grid construction.

\subsection*{Discussion}\label{sec12}
In this study, we propose a real-time power scheduling approach for a provincial power grid with high penetration of renewable energy. Our method is based on a planning-capable reinforcement learning algorithm and performs real-time rolling-horizon joint optimization. This eliminates the differences in the time scales between unit commitment and economic dispatch in traditional staged optimization, breaking away from the pre-calculation mode. This may solve the current dilemma of power scheduling, as our method no longer depends on the long-term renewable energy forecast results.
Our control design attains many of the expectations for a learning-based optimization approach in the power system community. It offers real-time computation, the ability to incorporate ultra-short-term forecasts, robustness to challenging scenarios, and anticipation of future events. These outcomes were made possible by bridging the gaps in capability and infrastructure through the integration of advances in reinforcement learning and electrical engineering. This includes the development of a realistic, accurate simulator and a security-constrained, planning-capable RL algorithm. The results of our real grid simulation demonstrate the potential to optimize more generators and larger power grids in real-time. RL-based methodology provides a promising direction for future power scheduling.

% Our learning framework has the potential to transform future research on power grid scheduling and operation. In the future, it could be used to explore configurations of storage management, demand response, or even the planning of electric vehicle charging. Additionally, it is also possible to use the AlphaZero-like Monte Carlo Tree Search (MCTS) to develop a warning mechanism that alerts operators of potential failures at an early stage. However, implementing this idea requires a server with hundreds of CPU cores, which cannot be achieved under our current experimental conditions. More broadly, our approach is compatible with an adversarial paradigm, in which the scheduling problem is treated as a board game between an attacker who attempts to collapse the power grid by attacking transmission lines and an operator who tries to make the power grid more stable by adjusting generators, etc. This could significantly improve the robustness of GridZero's policy. We hope to achieve these goals in future research.

% \section{Conclusion}\label{sec13}

\section*{Methods}\label{sec11}

In the following, we will describe the details of our proposed RL scheduling method.

\subsection*{GridSim environment}
In the GridSim environment, a provincial power grid with a high proportion of renewable energy is provided. The power grid consists of 126 buses, 185 transmission lines, 91 loads, and 54 generators. The power grid is described by an AC circuit model, and GridSim employs the power flow function to solve the coupled dynamics.

GridSim considers all the operational constraints that are present in real-world power scheduling scenarios.
In the case of generator startup/shutdown constraints, a thermal unit is allowed to be shutdown if reaching its minimum power in the last step, and then not allowed to be restarted in the next 40 steps, corresponding to the cooling process in the real world. If a thermal generator is restarted, it is not allowed to be shut down for the next 40 steps. 
For line outages, the disconnected lines are not allowed to be reconnected for 16 steps, corresponding to the maintenance period. 
Additionally, thermal generators have power ramping constraints due to their rotational inertia, which limits power changes within a single step. 

\subsubsection*{Reward design}
The reward function is crafted to provide guidance to the agent for compliance with the operational constraints and improvements on optimization objectives.
In accordance with the objectives and constraints of the RTS problem setting, the reward function is divided into distinct components.

\bmhead*{Line overflow.} 
To ensure the security of power transmission, we establish a reward mechanism that takes into account the current load rates of transmission lines. The current load rate is defined as the ratio of the transmission current to the maximum transmission current, which is limited by the thermal capacity of the line. 
To prevent line congestion and outages, it is desirable for the transmission lines to maintain a reasonable load level. Consequently, we design the reward function for this aspect as follows:
\begin{equation}
    r_{\text{overflow}}=1-\frac{\sum_i\min(\rho_i,1)}{n_{\text{line}}}
\end{equation}
where $\rho_i$ indicates the current load rate of line $i$. $n_{\text{line}}$ represents the line number. 

\bmhead*{Renewable consumption.} 
To optimize the utilization of renewable energy, a reward function is formulated to encourage the scheduling agent to consume more renewable energy. This reward function is based on the renewable energy consumption rate, which is defined as the ratio of the total power currently consumed from renewable sources to the maximum power generated by renewable sources. A higher renewable energy consumption rate corresponds to a higher reward, thereby incentivizing the agent to maximize the use of renewable energy. This reward is designed as follows:
\begin{equation}
    r_{\text{renewable}}=\frac{\sum_i\text{p}_i}{\sum_i\text{p}_i^{\text{max}}},\quad i\in\text{renewable units}
\end{equation}
where $\text{p}_i$ represents the power output of generator $i$, $\text{p}_i^\text{max}$ indicates maximal power generation capability of renewable generator $i$. The closer $\text{p}_i$ and $\text{p}_i^\text{max}$ are, the higher the renewable consumption reward.

\bmhead*{Balanced generator.} The balanced generator is to balance the residual power and eliminates discrepancies between power generation and load consumption, while its control capability is limited. If its power output exceeds its operational boundaries, a power mismatch occurs, which may result in load shedding or blackouts as previously mentioned. To prevent such failures and maintain safe operation, we design the reward function as follows:
\begin{gather}
    r_\text{balance} = -\left(\frac{\max(\text{p}_\text{bal}-\overline{\text{p}_\text{bal}},0)}{\overline{\text{p}_\text{bal}}-\underline{\text{p}_\text{bal}}}+\frac{\max(\underline{\text{p}_\text{bal}}-\text{p}_\text{bal},0)}{\overline{\text{p}_\text{bal}}-\underline{\text{p}_\text{bal}}}\right)
\end{gather}
where $\overline{p_\text{bal}}$ and $\underline{p_\text{bal}}$ indicate the upper bound and the lower bound of the balanced generator's active power. If the balanced power $p_\text{bal}$ is out of bounds, there would be a penalty.

\bmhead*{Operating cost.} We formulate a reward function for the operational costs of thermal units while considering the negligible costs of renewable energy generation. Specifically, the operating costs of thermal units are represented as quadratic functions of output power, and additional costs are incurred for the startup/shutdowns of thermal units. As for renewable sources, their operating costs are considered to be negligible as they do not rely on fossil fuels for power production. 
The operating cost reward is designed as follows:
\begin{equation}
    r_\text{cost} = -\frac{\sum_i c_{i,2}\text{p}_i^2+c_{i,1}\text{p}_i+c_{i,0}+\mathcal{I}(\text{s}_i, \text{s}_i^-)c_{\text{on-off,i}}}{Z}
\end{equation}
where $c_{i,2},c_{i,1}$ and $c_{i,0}$ are the second order, first order and constant coefficients of the operation cost of generator $i$, respectively. The coefficients of renewable units are much lower than that of thermal units. $\text{p}_i$ represents the power output of generator $i$. $\text{s}_i$ represents the on-off status of generator $i$, and the $\text{s}_i^-$ is the status 1-step advance. $c_\text{on-off,i}$ is the startup and shutdown costs of generator $i$. $\mathcal{I}(\text{s}_i, \text{s}_i^-)$ is an indicative function that turns to be 1 if $\text{s}_i\neq\text{s}_i^-$, otherwise 0. $Z$ is the normalization factor set as $10^5$ in experiments.

\bmhead*{Reactive power.} Reactive power plays a vital role in supporting the voltage stability of the power grid. However, the reactive power output capacity of the generators is constrained. While exceeding this limit is not catastrophic, excessive reactive power compensation can significantly increase operational cost. In light of these considerations, we design 
the reactive power reward as follows:
\begin{equation}
    r_\text{reactive}=\exp\left(-\sum_i\left[\frac{\max(\text{q}_i-\overline{\text{q}_i},0)}{\overline{\text{q}_i}-\underline{\text{q}_i}}+\frac{\max(\underline{\text{q}_i}-\text{q}_i,0)}{\overline{\text{q}_i}-\underline{\text{q}_i}}\right]\right)-1
\end{equation}
where $\text{q}_i$ is the reactive power of generator $i$, and $\overline{\text{q}_i}, \underline{\text{q}_i}$ are the upper bound and the lower bound of generator $i$. There would be a penalty if any generator violates its reactive power constraint. 

\bmhead*{Bus voltage.} In power system operation, it is common practice to limit node voltage magnitudes within the range of 0.95-1.05 per unit. If the voltage magnitude at a node is too low, it can result in a significant increase in the transmission loss of the grid. Conversely, if the node voltage magnitude is too high, it requires more reactive power compensation and may cause the generator's reactive power to exceed its upper limit.
To regulate the node voltage magnitudes within specified ranges, we design the bus voltage reward similarly to the reactive power reward.
\begin{equation}
    r_{\text{voltage}}=\exp\left(-\sum_i\left[\frac{\max(\text{v}_i-\overline{\text{v}_i},0)}{\overline{\text{v}_i}-\underline{\text{v}_i}}+\frac{\max(\underline{\text{v}_i}-\text{v}_i,0)}{\overline{\text{v}_i}-\underline{\text{v}_i}}\right]\right)-1
\end{equation}
where $\text{v}_i$ is the voltage magnitude of bus $i$, and $\overline{\text{v}_i}, \underline{\text{v}_i}$ are the upper bound and the lower bound of voltage magnitude of bus $i$. There would be a penalty if any bus violates its voltage magnitude constraint. 

The total reward is calculated as the weighted sum of the above reward parts.

\subsection*{GridZero}
\begin{figure}[h]
  \centering
  \includegraphics[width=1.0\linewidth]{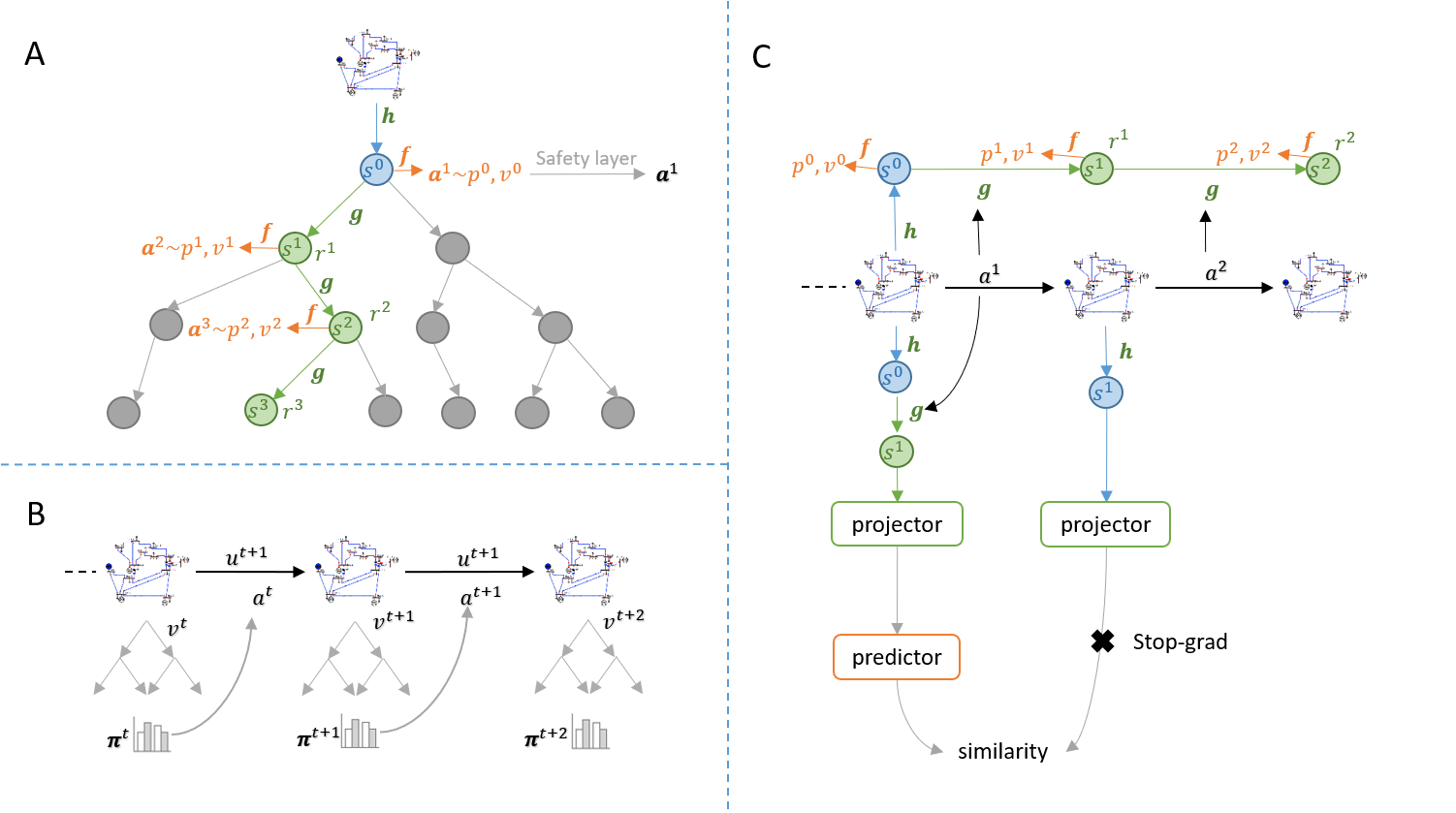}
  \caption{\textbf{Planning, acting, and training of GridZero.} \textbf{(A) Planning.} The model includes representation $h$, dynamics $g$, and prediction function $f$. The dynamics network $g$ predicts the reward $r^t$ and the next hidden state $s^t$ when given a hidden state $s^{t-1}$ and action $a^t$. The prediction network $f$ computes the policy $p^t$ and value estimation $v^t$ using $s^t$. Action candidates $\{a^t_i\}$ are sampled from $p^t$, with root action candidates processed by the safety layer. The representation network $h$ embeds the hidden state $s^t$ using the observation of power grid $o^t$. \textbf{(B) Acting.} An MCTS tree is expanded at each step $t$. An original action $a^{t+1}$ is sampled from the visit count distribution $\pi^t$, mapped to a real action $\tilde{a}^{t+1}$ by the action mapping layer, and then used to simulate a new observation $o^{t+1}$ and return a ground-truth reward $u^{t+1}$. \textbf{(C) Training.} The model is unrolled for $K$ steps.
  At each step $k$, the dynamics network $g$ receives $s^{k-1}$ and the action $a^{k}$ as input, generates the next hidden state $s^{k}$ and reward estimation $r^{k}$. GridZero learns by mimicing the policy $p^{k}\to\pi^{k}$, value estimation $v^{k}\to z^{k}$, and reward estimation $r^{k}\to u^{k}$, where $z^{k}$ is the bootstrap prefix value target. GridZero also introduces a consistency loss by using the self-supervised network to calculate the similarity between the estimated hidden state $\hat{s}^{k+1}$ and the target hidden state $s^{k+1}$.}
  \label{fig:how_gridzero_work}
\end{figure}

Complex action spaces are common indeed in real-world problems, and so is the real-time scheduling (RTS) problem. The RTS problem requires the RL agent to make decisions simultaneously on optimal power outputs and generators' startup/shutdown. 
This section outlines the combination of Sampled MuZero~\cite{hubert2021learning} and EfficientZero~\cite{ye2021mastering} to develop the core GridZero algorithm for solving the online control of RTS.  
To provide a better understanding of our work, we use Fig.\ref{fig:how_gridzero_work} to demonstrate how GridZero performs planning, acting, and training.

We first present the sampling method and the hybrid policy, which allows GridZero to accommodate to the hybrid action space of power scheduling, consisting of both discrete startup/shutdowns and continuous power outputs. 
Action candidates for MCTS are sampled from the hybrid policy $p$, which includes a continuous power adjustment vector and startup/shutdown one-hot vectors. 
In MCTS, the agent selects the action of the child with with the highest Upper Confidence Bound (UCB) score to visit, which is formulated as
\begin{equation}
    a=\mathop{\arg\max}\limits_{a\sim p}Q(s,a)+\hat{\beta}(s,a)\frac{\sqrt{\sum_b N(s,b)}}{1+N(s,a)}
\end{equation}
The hybrid policy $p$ consists of a Gaussian distribution corresponding to the continuous policy and a categorial distribution corresponding to the discrete policy. $Q(s,a)$ and $N(s,a)$ is the Q-value estimation and visit count of state-action pair $(s,a)$. $\hat{\beta}(s,a)$ is a uniform prior distribution in GridZero's setting. 

In order to train GridZero, the loss function is formulated as
\begin{equation}
    l=\sum_{k=0}^K l^r(u^{k},r^{k})+l^v(z^{k},v^{k})+l^p(\pi^{k},p^{k},\bm{a}^k)+l^c(s^{k+1},\hat{s}^{k+1})+H(p^{k})
    % +c\Vert\theta\Vert^2
\end{equation}
The loss function is computed over a horizon of $K$ unrolled steps. The reward loss $l^r$ measures the difference between the estimated reward $r^k$ and target reward $u^{k}$. Similarly, The value loss $l^v$ indicates the difference between the estimated value $v^k$ and bootstrapped target value $z^{k}$. The policy loss $l^p$ represents the distance between the output policy $p^k$ and the target root visit distribution of MCTS $\pi^{k}$, and $\bm{a}^k$ is the root action candidates at step $k$.
In order to address the challenge of insufficient supervisory signal in dynamic networks in MuZero, we introduce consistency loss $l^c$ which maximizes the similarity between the predicted next-state $\hat{s}^{k+1}$ and the target next-state $s^{k+1}$ derived from the observation $h(o^{k+1})$. Additionally, we incorporate the entropy loss term $H(p^{k})$, following the SAC's settings, to encourage exploration in the learning process\cite{haarnoja2018soft}. 

\subsubsection*{Hybrid Action Sampling}
To more accurately represent the power output adjustment and switching behaviors of generators, we utilize a hybrid action that concatenates a continuous vector in the range of $[-1,1]^n$ with two one-hot vectors $\left\{0,1\right\}^{n+1}$, representing the control of active power and startups/shutdowns of generators, respectively.

In brief, the output of the policy network is now split into four parts: Gaussian means $p_\mu$, Gaussian logarithmic standard deviations $p_\sigma$, discrete startup logits $p_o$ and discrete shutdown logits $p_c$. The dimensions of $p_o$ and $p_c$ are both $n+1$, where $n$ is the number of generators, with the additional dimension corresponding to no startup or shutdown. The active power control vector $a_p$ is generated by sampling from a squashed multivariate Gaussian distribution $\overline{\mathcal{N}}(p_\mu, p_\sigma)$. The one-hot vectors of startups $a_o$ and shutdowns $a_c$ are sampled from categorial distributions $p_o$ and $p_c$. An action candidate $a_i$ of the root is the concatenation of $a_p$, $a_o$ and $a_c$. The candidates of root nodes are required to be corrected by the safety layer since random exploration is not allowed in this security-constrained task. Our action mapping layer only maps the selected root action candidate $a$ to a practical action as follows:
\begin{equation}
    \tilde{a} = 
    \frac{a_\text{p}-(-1)}{1-(-1)}
    % \frac{a_\text{p}+1}{2}
    \cdot(\overline{\text{p}}-\underline{\text{p}})+\underline{\text{p}}+\overline{\text{p}}\cdot a_o+\underline{\text{p}}\cdot a_c
\end{equation}
where $\overline{\text{p}}, \underline{\text{p}}$ are the time-varying upper bound and lower bound of the action space, $[a_p, a_o, a_c]$ is the original sampled action, including active power control, startup/shutdown one-hot vectors.

\subsubsection*{Design of Power Balancing Safety Layer}
Exploration without violations to constraints is a critical challenge in our problem. This is because not all states are accessible, particularly those that may result in severe consequences. 
The RTS problem has complex operational constraints, however, the most crucial constraint is power balancing, which requires that the power generation be equal to load consumption at any time.
The fluctuating loads and renewable generation introduce temporal variations into the power balancing constraint, which makes it more difficult to maintain this constraint.
We found that simple reward shaping and auxiliary loss methods cannot solve this problem since the action space is high-dimensional and time-varying. 

Therefore, We propose a safety layer to ensure that the agent satisfies this basic power balancing constraint, which is equivalent to restricting the feasible solution space. 
The safety layer calculates the readjustment power $\Delta a$ through two objectives, balancing the predicted change in total load $\Delta\hat{\text{p}}_{\text{load}}=\hat{\text{p}}^{t+1}_\text{load}-\text{p}^t_\text{load}$, and 'pulling' the deviated balanced generator power back to its safe range $[\underline{\text{p}_{\text{bal}}}+\delta, \overline{\text{p}_{\text{bal}}}-\delta]$.
With this information, we calculate a readjustment objective $\Delta a$ for other generators using the following equation:
\begin{equation}
\label{eq:safety_obj}
    \Delta a=\Delta\hat{\text{p}}_{\text{load}}-\sum_i \tilde{a}_i+\Delta \text{p}_{\text{bal}}
\end{equation}
where 
\begin{equation}
    \label{eq:safety_pbal}
    \Delta \text{p}_{\text{bal}}=
    \left\{
             \begin{array}{lr}
             -(\overline{\text{p}_{\text{bal}}}-\delta-\text{p}_{\text{bal}}), & if\ \text{p}_{\text{bal}}>\overline{\text{p}_{\text{bal}}}-\delta \\
             \text{p}_{\text{bal}}-\underline{\text{p}_{\text{bal}}}-\delta, & if\ \text{p}_{\text{bal}}<\underline{\text{p}_{\text{bal}}}+\delta. 
             \end{array}
\right.
\end{equation}
with $\delta$ as the redundancy to the limits. $a_i$ refers to the power adjustment of generator $i$. 
% Once the overall readjustment target $\Delta a$ is derived, we calculate the adjustment for each unit to form the feasible action. 
The readjustment target $\Delta a$ will be allocated to each online generator in proportion. The allocation proportion is determined by the generator's climbing capacity. The lager the climbing capacity, the greater the allocation power.
The closed generators are supposed to be masked out because they are not adjustable. We only readjust those adjustable units if the objective $\Delta a$ exceeds a threshold, which means that the safety layer is not always active. This is because we want to enable the agent reasonable exploration space, as long as it is within the power balancing constraint.  In our experiments, the threshold is a number smaller than the redundancy $\delta$. 
We can calculate the readjustment of each generator as follows:
\begin{equation}
\label{eq:safety_readj_value}
    \Delta a_{i}=
    \left\{
             \begin{array}{lr}
             \Delta a\cdot(\overline{\text{p}_i}-a_i)/\sum_i(\overline{\text{p}_i}-a_i), & if\ \Delta a>0 \\
             \Delta a\cdot(a_i-\underline{\text{p}_i})/\sum_i(a_i-\underline{\text{p}_i}), & if\ \Delta a\leq 0. 
             \end{array}
\right.
\end{equation}
so the legalized action turns to be $a_i=a_i+\Delta a_i$. 

Such a safety layer essentially maps the original action space of the policy network output to a time-varying feasible solution space, which satisfies the fundamental power balancing constraint. 

\subsubsection*{Exploration noises on root nodes}

To ensure enough exploration to approach a near-optimal control policy, we propose two types of noises for both continuous active power control and discrete switching actions. 
For discrete actions, we introduce Dirichlet noises to the switching probabilities, resulting in the one-hot vectors being sampled from the perturbed categorical distribution. Regarding continuous actions, we consider two approaches. The first approach is to add standard normal noises to the sampled actions, which can result in more radical explorations. The second approach involves sampling from a more flattened Gaussian by doubling the variance, as in $\bar{\mathcal{N}}(p_\mu, kp_\sigma)$, thereby leading to conservative explorations around the current policy. Finally, we use a mixture of the above methods:
% First, we add Dirichlet noises to the switching probabilities for discrete actions, and then the one-hot vectors are sampled from the disturbed categorial distribution. One approach for continuous actions is adding standard normal noises to sampled actions, which might produce more radical explorations. Another approach is to sample from a more flattened Gaussian by doubling the variance like $\bar{\mathcal{N}}(p_\mu, kp_\sigma)$, leading to conservative explorations in the vicinity of the current policy. Finally, we use a mixture of the above as follows:
\begin{equation}
    a=
    \left\{
             \begin{array}{lr}
             \sim\mathcal{N}(p_\mu, p_\sigma), & \text{sample $K_{\text{normal}}$ actions} \\ 
             \sim\mathcal{N}(p_\mu, p_\sigma)+\mathcal{N}(0,1), & \text{sample $K_{\text{bigger}}$ actions} \\
             \sim\mathcal{N}(p_\mu, kp_\sigma), & \text{sample $K_{\text{smaller}}$ actions}
             \end{array}
    \right.
\end{equation}
where total action candidates number is $K=K_{\text{normal}}+K_{\text{bigger}}+K_{\text{smaller}}$. In self-play, all the disturbed sampled actions of the root node should be processed by the safety layer.

\subsubsection*{Policy loss for hybrid actions}
To suitably tailor GridZero to accommodate a complex action space, we have implemented certain alterations to GridZero's policy loss. 
Within our hybrid action sampling, an action candidate $a$ is comprised of concatenated power adjustments $a_p$, startup binary one-hot encoding $a_o$, and shutdown binary one-hot encoding $a_c$. Concerning the continuous aspect of this framework, the loss function is represented as the Kullback-Leibler(KL) divergence of root distribution and policy distribution, as exemplified in Eq.\ref{eq:policy_loss_continuous}.
\begin{equation}
\label{eq:policy_loss_continuous}
    l^p_c=-\pi^\top\log p(a_p)
\end{equation}
where $a_p$ is the action candidate sampled from Gaussian policy $p(\cdot)=\mathcal{N}(p_\mu,p_\sigma)$. $\pi$ is the visit count distribution of the MCTS root node. In the case of discrete actions, we sample a limited number of actions because there are numerous possible combinations of generator switches. If all the combinations were taken into account, it would affect the search efficiency of MCTS. Nonetheless, this will introduce a dimension mismatch between the discrete policy distribution and the root visit distribution. To address this mismatch, we perform dot production of visit priors and one-hot actions as $\pi\cdot a_o$, which serves as the target switching policy. The loss function for the discrete policy loss is given by:
\begin{equation}
    l^p_d=-(\pi\cdot a_o)^\top \cdot\text{LogSoftmax}(p_o)-(\pi\cdot a_c)^\top\cdot\text{LogSoftmax}(p_c)
\end{equation}
where $p_o$ and $p_c$ are policy logits for units' startup/shutdown. $a_o$ and $a_c$ are target startup and shutdown actions. The total policy loss is a combination of continuous part and discrete part:
\begin{equation}
    l^p=l_c^p+l_d^p
\end{equation}

\backmatter

% \bmhead{Supplementary information}

% If your article has accompanying supplementary file/s please state so here. 

% Authors reporting data from electrophoretic gels and blots should supply the full unprocessed scans for key as part of their Supplementary information. This may be requested by the editorial team/s if it is missing.

% Please refer to Journal-level guidance for any specific requirements.

\bmhead{Acknowledgments}

This work is supported by the Ministry of Science and Technology of the People´s Republic of China, the 2030 Innovation Megaprojects "Program on New Generation Artificial Intelligence" (Grant No. 2021AAA0150000) and the National Natural Science Foundation of China (Grant No. 52122706).

\bmhead{Authors' contribution}
Shaohuai Liu, Weirui Ye, Yang Gao conceived the idea of GridZero. Shaohuai Liu developed the GridZero's code and wrote the manuscript. Chongqing Kang, Haiwang Zhong and Guanglun Zhang provided revision for power system part. Qirong Jiang and Yang Gao provided funding and revision. Jinbo Liu accomplished the architecture design and scheme formulation of GridSim. Fangchun Di accomplished grid model design and data generation of GridSim. Yang Nan finished the code development of GridSim. 
% Please refer to Journal-level guidance for any specific requirements.

\bmhead{Competing interests}
The authors declare no competing interests.

\bmhead{Code availability}

The reinforcement learning code is available at \url{https://github.com/liushaohuai5/GridZero.git}.

\bmhead{Data availability} GridSim simulator is limited open-sourced. If required, contact CEPRI for permission.

\bibliography{sn-bibliography}% common bib file

\newpage
\subsection*{Supplementary information [Not applicable]}

\subsubsection*{Observation formulation of GridSim}
As shown in Extended Table.1, the observation is a dictionary of different values, including time, generator states, load states, line status and ultra-short-term forecasts.
\begin{table}[h]
    \centering
    \begin{tiny}
    \begin{tabular}{ccc}
    \toprule
        Variable Name & Data Type & Meaning \\
    \midrule
        timestep & int & current steps \\
        vTime & string & current timestamp \\
        gen\_p & list[float] & active power of generators \\
        gen\_q & list[float] & reactive power of generators \\
        gen\_v & list[float] & voltage magnitudes of generators \\
        load\_p & list[float] & active power of loads \\
        load\_q & list[float] & reactive power of loads \\
        load\_v & list[float] & voltage magnitudes of loads \\
        line\_status & list[bool] & line status \\
        grid\_loss & float & transmission loss \\
        action\_space & dict & legal action space of the next step\\
        rho & list[float] & line load rate\\
        gen\_status & list[bool] & generator status, 1-on/0-off \\
        steps\_to\_recover\_gen & np.ndarray & number of steps for closed units to restart \\
        steps\_to\_close\_gen & np.ndarray & number of steps for restarted units to shutdown \\
        curstep\_renewable\_gen\_p\_max & list[float] & max power of the renewable unit at the current step \\
        nextstep\_renewable\_gen\_p\_max & list[float] & max power of the renewable unit at the next step \\
        next\_load\_p & list[float] & load for the next step \\
    \bottomrule
    \end{tabular}
    \captionsetup{labelformat=empty}
    \caption{\textbf{Extended Table. 1:} Main components of $GridSim.observation$. Please refer to the supplementary data for more details.}
    \end{tiny}
    \label{tab:observation}
\end{table}

\subsubsection*{Action space formulation}
As shown in Extended Table.2, the action space is a dictionary of the upper boundaries and lower boundaries of generators' output power. For thermal generators, their upper bounds and lower bounds are determined as $[-\text{p}_\text{ramp}, \text{p}_\text{ramp}]$. For renewable generators, since their output power could be set as $[0,\overline{\text{p}}]$ at each step, their upper bounds and lower bounds are $[-\text{p}, \overline{\text{p}}-\text{p}]$.
\begin{table}[h]
    \centering
    \begin{tiny}
    \begin{tabular}{ccc}
    \toprule
        Variable Name & Data Type & Meaning \\
    \midrule
        % adjust\_gen\_p & np.ndarray & active power adjustments of generators \\
        adjust\_gen\_p.high & np.ndarray & upper bound of active power adjustment of generators\\
        adjust\_gen\_p.low & np.ndarray & lower bound of active power adjustment of generators\\
    \bottomrule
    \end{tabular}
    \captionsetup{labelformat=empty}
    \caption{\textbf{Extended Table. 2:} Main components of $GridSim.action\_space$. Please refer to the supplementary data for more details.}
    \end{tiny}
    \label{tab:observation}
\end{table}

\subsubsection*{Static parameters of GridSim}
Some static parameters of GridSim are shown in Extended Table.3, including the indexes of renewable generators, power ramping rate of thermal generators, etc. 
\begin{table}[h]
    \centering
    \begin{tiny}
    \begin{tabular}{ccc}
    \toprule
        Variable Name & Data Type & Meaning \\
    \midrule
        num\_gen & int & number of generators \\
        num\_line & int & number of lines \\ 
        num\_load & int & number of loads \\
        num\_bus & int & number of buses \\
        gen\_type & list[int] & generator types, 5-renewable, 1-thermal, 2-balanced\\
        max\_gen\_p & list[float] & upper bound of active power output \\
        min\_gen\_p & list[float] & lower bound of active power output \\
        max\_gen\_q & list[float] & upper bound of reactive power output \\
        min\_gen\_q & list[float] & lower bound of reactive power output \\
        max\_gen\_v & list[float] & upper bound of generator voltage magnitude \\
        min\_gen\_v & list[float] & lower bound of generator voltage magnitude \\
        ramp\_rate & float & active power ramping factor of thermal generators \\
        thermal\_ids & list[int] & thermal generators index \\
        renewable\_ids & list[int] & renewable generators index \\
        balanced\_id & list[int] & balanced generator index\\
        startup\_cost & list[float] & starup cost\\
        constant\_cost & list[float] & constant cost \\
        first\_order\_cost & list[float] & first order cost coefficient \\
        second\_order\_cost & list[float] & second order cost coefficient\\
    \bottomrule
    \end{tabular}
    \captionsetup{labelformat=empty}
    \caption{\textbf{Extended Table. 3:} Main components of $GridSim.static\_parameters$. Please refer to the supplementary data for more details.}
    \end{tiny}
    \label{tab:observation}
\end{table}

\subsubsection*{Rules of GridSim}
\paragraph{The upper and lower limits of the active output of the unit} The active power of any unit (except the balancing unit) cannot be greater than the upper limit of the active power, nor can it be smaller than the lower limit of the active power. If it is violated, the emulator prompts "illegal action", forcibly ending the episode.

\paragraph{Maximum output constraint of new energy units} 
In any time step, the active power output of renewable energy units cannot be greater than the maximum power generation capability. If it is violated, the emulator prompts "illegal action", forcibly ending the round.

\paragraph{Unit ramping constraint} The active power adjustment of any thermal power unit must be smaller than the ramping rate. If it is violated, the emulator prompts "illegal action", forcibly ending the episode.

\paragraph{Unit startup/shutdown constraints} The shutdown rule for thermal power units is that the active power output of the unit must be adjusted to the lower bound before the unit is shutdown, and then adjusted to 0. Restarting is not allowed within 40 consecutive steps after the unit is stopped. The startup rule of thermal power units is that the active power output must be adjusted to the lower bound before the unit is turned on. No shutdown is allowed for 40 consecutive time steps after the unit is restarted.

\paragraph{Line overflow constraint} If the line current exceeds the limit but does not exceed 135\% of the thermal limit, it means that the branch is "soft overloaded". If the line current exceeds 135\% of the thermal regulation limit, it means the line is "hard overloaded". If any line has "soft overload" for 4 consecutive time steps, the line will be shut down. In the event of a "hard overload", the line is immediately shut down. After the line has been out of service for 16 time steps, it will be put into operation again.

\paragraph{Upper and lower limit constraints of the unit's reactive power output} When the agent adjusts the voltage at the machine terminal, the unit's reactive power output value exceeds its upper and lower limits, and a negative reward will be obtained.
\paragraph{Voltage upper and lower limit constraints} If the node voltage exceeds its upper and lower limits, it will get a negative reward.

\paragraph{Upper and lower limit constraints of the balancing machine} The system sets a balancing unit to share the unbalanced power of the system caused by the unreasonable control strategy. After the power flow calculation, if the active power output of the balancing unit is greater than the upper limit but less than 110\% of the upper limit, or less than the lower limit but greater than 90\% of the lower limit, a negative reward will be obtained. If the output is greater than 110\% of the upper limit or less than 90\% of the lower limit, the episode ends.

\subsubsection*{Training statistics}
As shown in Extended Fig.1, GridZero much outperforms SAC, which shows the superiority of planning over the actor-critic framework in terms of policy improvement. All experiments are done on a machine with 8 Nvidia RTX-2080Ti GPUs and 80 cores of Intel Xeon Gold 5218R CPU.

\begin{figure}[h]
\centering
% \begin{minipage}[t]{0.6\linewidth}
% \centering
\includegraphics[width=0.6\linewidth]{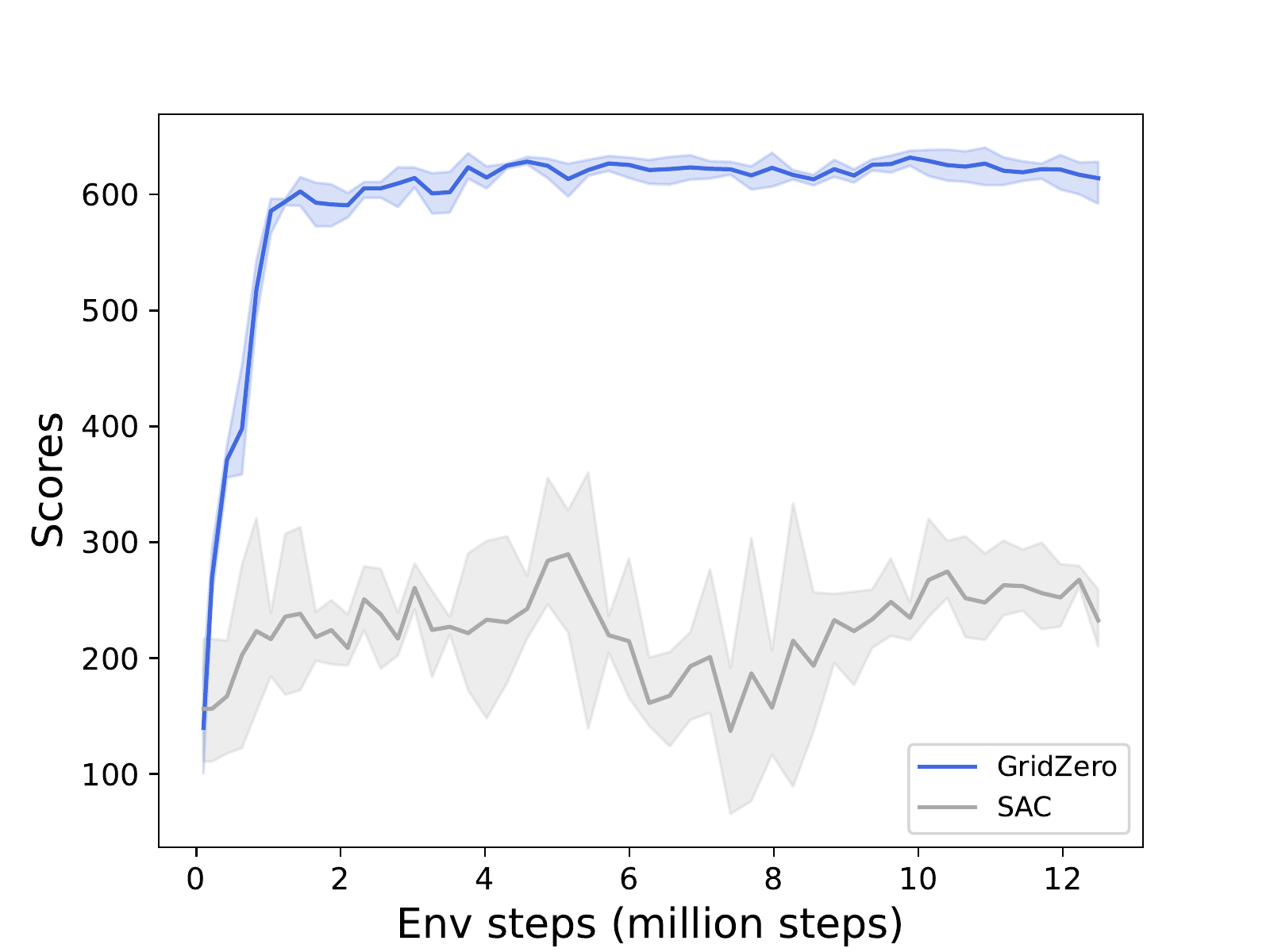}
\captionsetup{labelformat=empty}
% \end{minipage}
\caption{\textbf{Extended Fig. 1:}
Performance curves of GridZero and SAC. The X-axis represents environmental steps. The Y-axis indicates the average episode scores. GridZero much outperforms the model-free baseline SAC. The average scores among 20 evaluation seeds for 4 runs is shown on the Y-axis. }
\label{fig:perf-sns-trade-off}
\end{figure}

Moreover, Extended Fig.1 and Extended Fig.2 both show that GridZero is highly sample-efficient that it only requires 1 million environmental interaction steps and 15-kilo training steps.
\begin{figure}[h]
    \centering
    \includegraphics[width=0.6\linewidth]{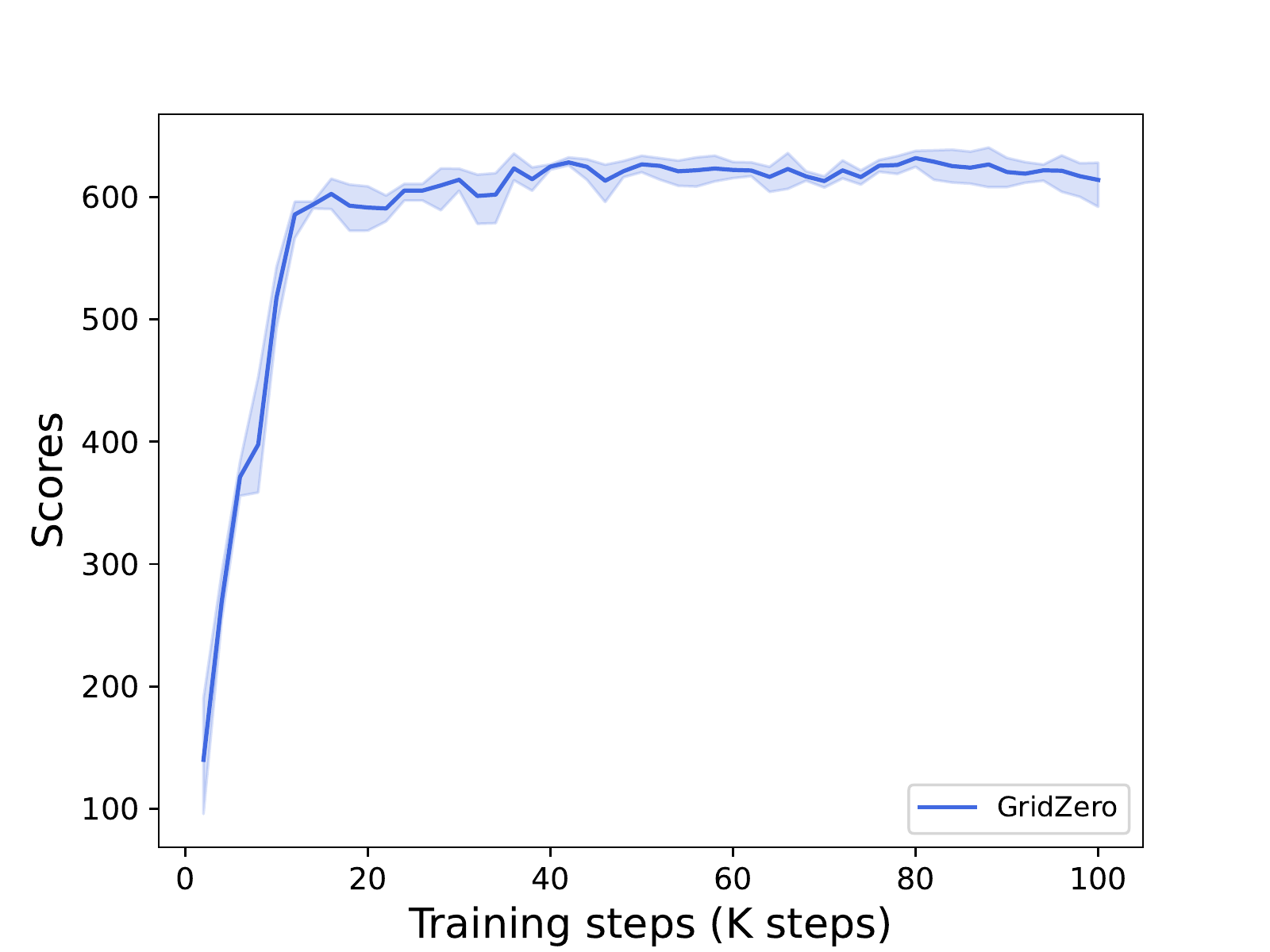}
    \captionsetup{labelformat=empty}
    \caption{\textbf{Extended Fig. 2:} Training curves of GridZero. X-axis represents training steps. Y-axis indicates episode scores. The average scores among 20 evaluation seeds for 4 runs is shown on the Y-axis.}
    \label{fig:perf_training}
\end{figure}

Extended Fig.3 shows GridZero's performance in 20 different scenarios. In each scenario, GridZero significantly reduces renewable curtailment, and keeps the adjustment capacity area well covering the load curve.
\begin{figure}[h]
  \centering
  \includegraphics[width=1.0\linewidth]{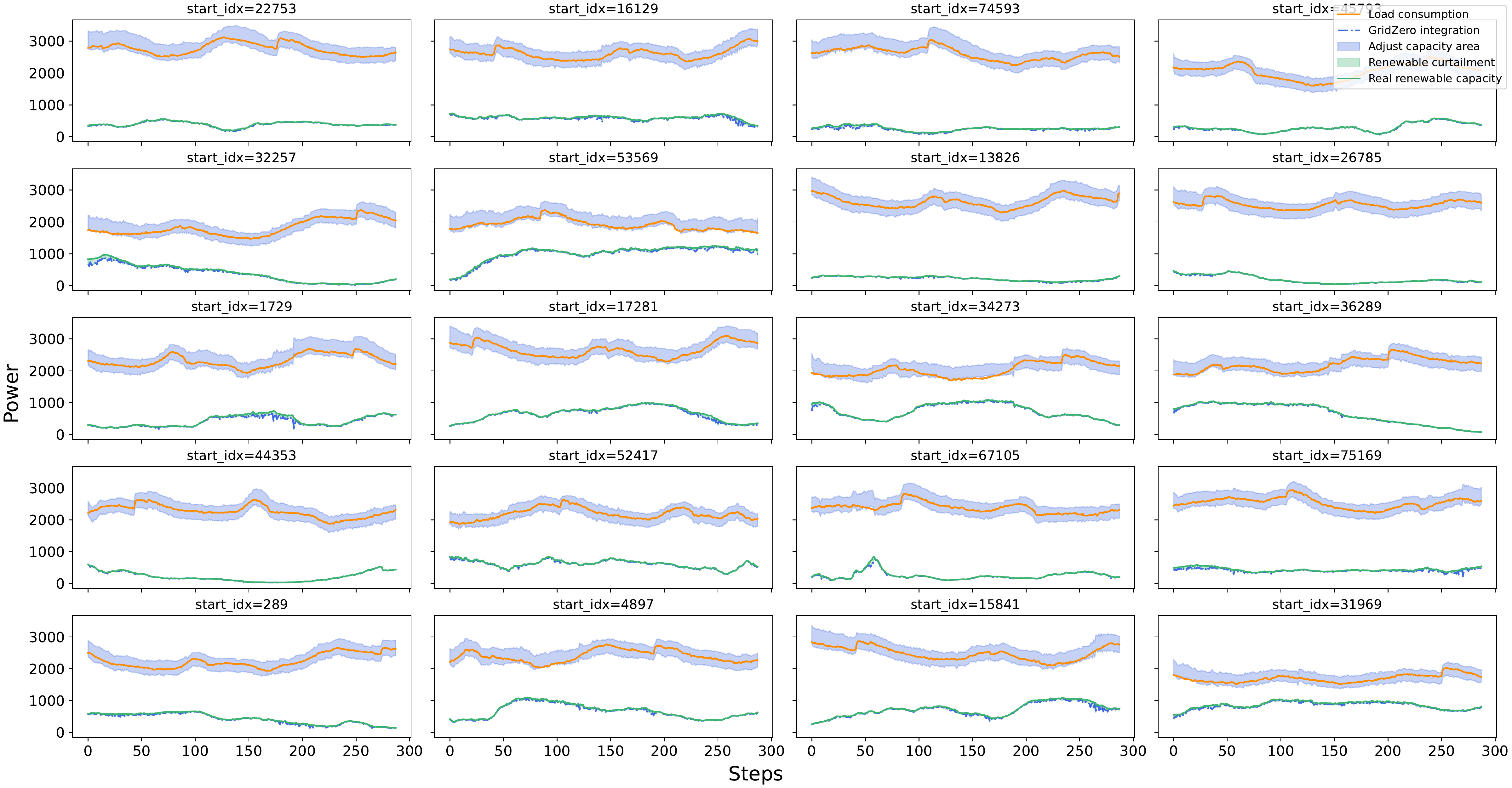}
  \captionsetup{labelformat=empty}
  \caption{\textbf{Extended Fig. 3:} Renewable consumption and load curves of 20 test scenarios. X-axis represents 288 dispatching steps over a whole day. Y-axis indicates power. GridZero could well track maximum power curve of renewable sources and keep the adjust capacity area covering the load consumption curve. 
  % In renewable energy rich scenarios, such as section 53569, GridZero can adjust the on-off status of thermal units to bring as much "green electricity" onto the grid as possible. GridZero achieves similar performance to AC-OPF in renewable energy consumption. The early plunge to 0 of curves of SAC and DDPG means that those model-free agents caonnot run the whole episode.
  } 
  \label{fig:20_episodes}
\end{figure}

Extended Fig.4 shows the learning process of different constraints, the operation cost and the renewable consumption rate. All constraint violation rates decrease with the training process, also including the operation cost. GridZero also achieves 95\% of an average renewable consumption rate.
\begin{figure}[h]
    \centering
    \includegraphics[width=1.0\linewidth]{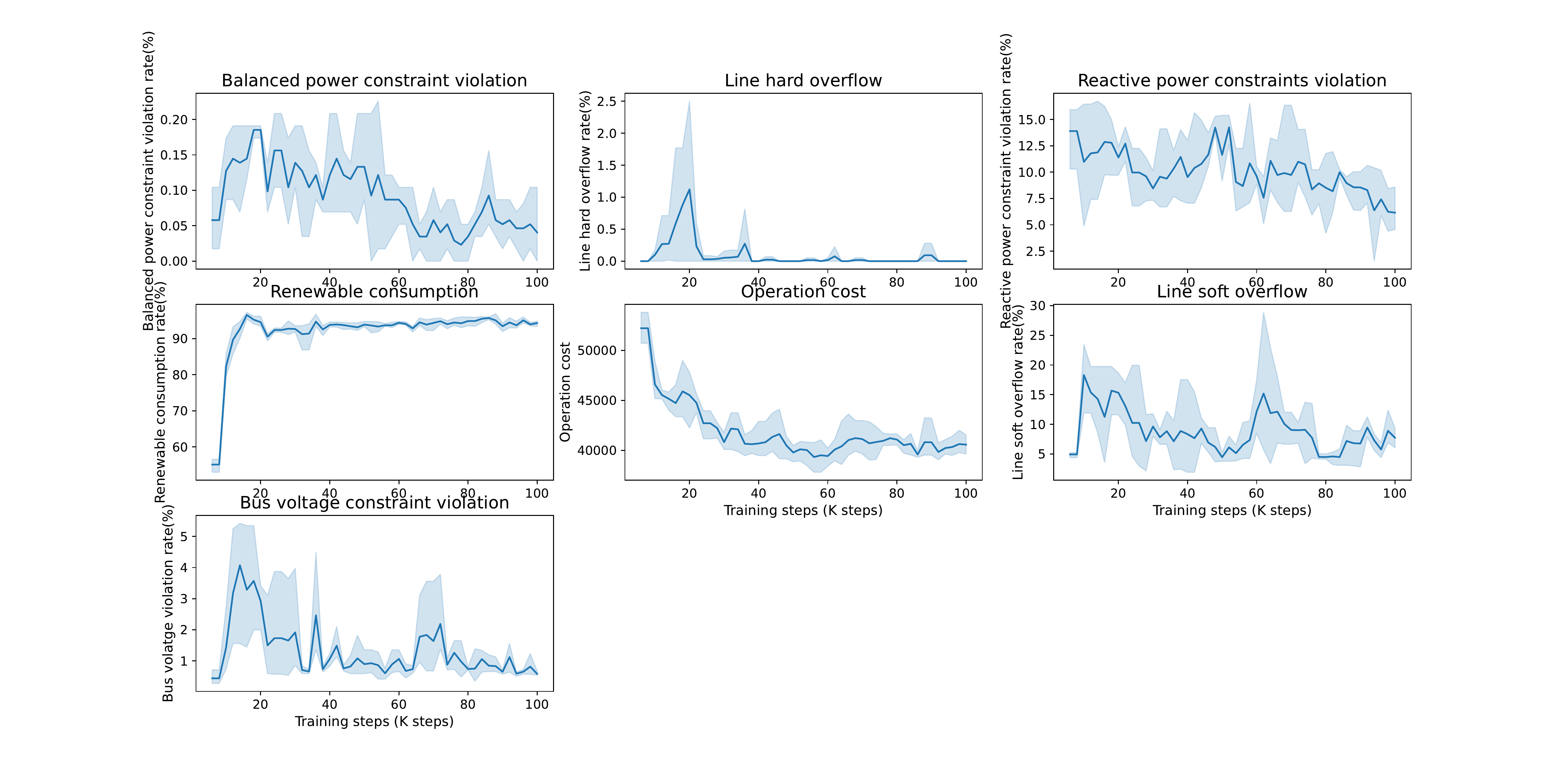}
    \captionsetup{labelformat=empty}
    \caption{\textbf{Extended Fig. 4:} Learning curves of constraints and objectives. X-axis represents the training steps. Y-axis represents the violation rates or objective values.}
    \label{fig:my_label}
\end{figure}

\subsubsection*{Network architecture}
As shown in  Extended Fig.5, since RTS is a state-based task. All of our networks are implemented by simple Multi-Layer Perceptron (MLP). The more complex network architectures, such as Graph Neural Networks (GNN), could be further studied in future works.
\begin{figure}[h]
    \centering
    \includegraphics[width=1.0\linewidth]{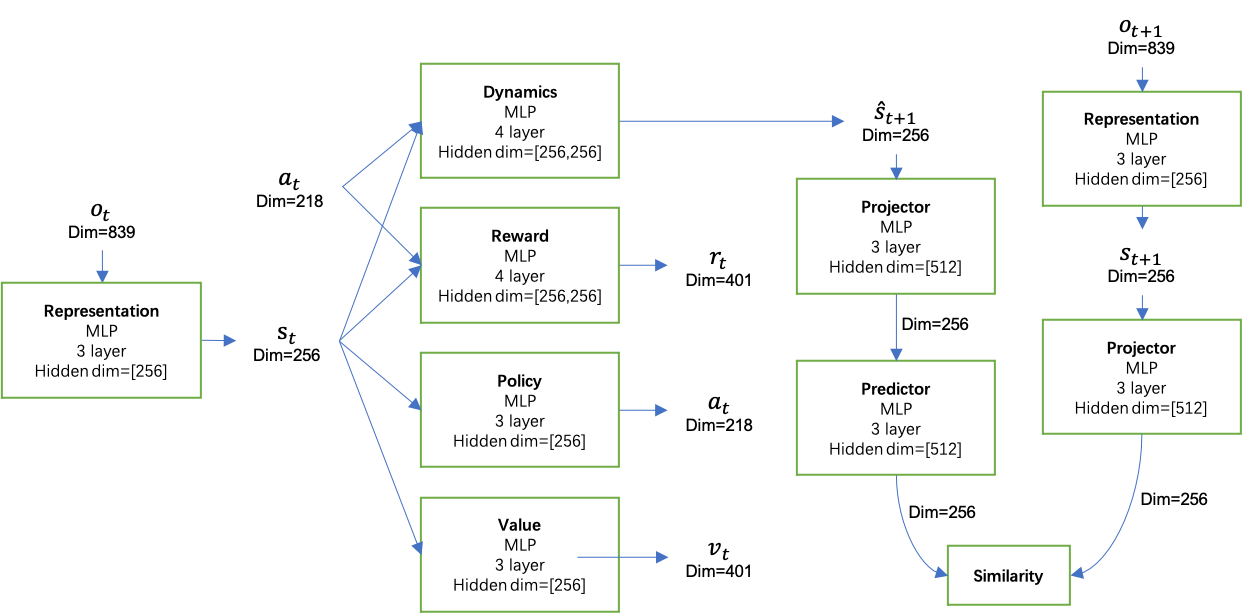}
    \captionsetup{labelformat=empty}
    \caption{\textbf{Extended Fig. 5:} Network architecture of GirdZero. All sub-networks are MLPs.}
    \label{fig:net_arch}
\end{figure}

\subsubsection*{Power Grid Architecture}
The power grid architecture is shown in Extended Fig.6, including 126 buses, 185 lines, 54 generators, and 91 loads. This provincial grid is divided into three sub-networks, scattered in different areas.
\begin{figure}[h]
    \centering
    \includegraphics[width=1.0\linewidth]{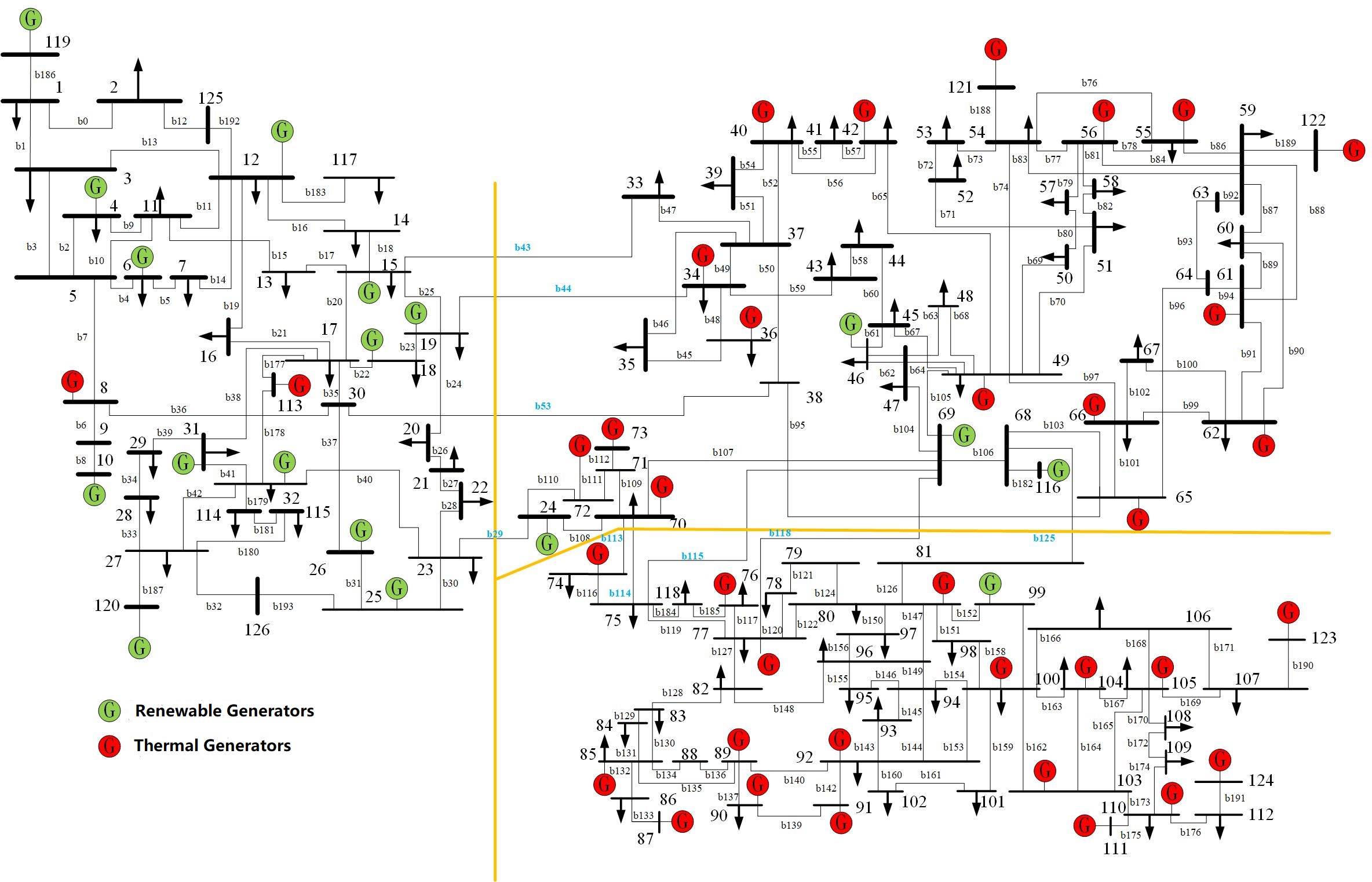}
    \captionsetup{labelformat=empty}
    \caption{\textbf{Extended Fig. 6:} Provincial power grid of GridSim, includes 126 buses, 185 lines, and 54 generators, 17 of which are renewable.}
    \label{fig:grid_arch}
\end{figure}

\subsubsection*{Hyperparameters}
As shown in Extended Table.4, we released some hyperparameters for training settings to help researchers who are interested in GridZero better reproduce our work.
\begin{table}[h]
    \label{tab:param}
\begin{center}
\begin{small}
% \begin{sc}
\centering
\scalebox{1.0}{
\centering
\begin{tabular}{lc}
\toprule
Parameter & Setting \\
\midrule
Observation & 839 \\
Frames stacked & 1 \\
Frames skip & 0 \\
Reward clipping & True \\
Reward clipping delta & 0.1 \\
Max frames per episode & 288 \\
Discount factor & 0.99 \\
Minibatch size & 256 \\
Optimizer & SGD \\
Optimizer: learning rate & 0.01 \\
Optimizer: momentum & 0.9 \\
Optimizer: weight decay ($c$) & 2e-5 \\
Learning rate schedule & exponential, 0.5 \\
Max gradient norm & 10 \\
Priority exponent ($\alpha$) & 0.6 \\
Priority correction ($\beta$) & 0.4 \\
Training steps & 100k \\
Self-play network updating inerval & 100 \\ 
Target network interval & 200 \\
Unroll steps ($l_{\text{unroll}}$) & 5 \\
TD steps ($k$) & 5 \\
Policy loss coefficient ($\lambda_1$) & 1 \\
Value loss coefficient ($\lambda_2$) & 0.5 \\
Reward loss coefficient ($\lambda_3$) & 1.0 \\
Self-supervised consistency loss coefficient ($\lambda_4$) & 2.0 \\
Dirichlet noise alpha ($\xi$) & 0.3 \\
Dirichlet noise ratio & 0.25 \\
Number of simulations in MCTS ($N_{\text{sim}}$) & 50 \\
Number of sampled normal actions & 13 \\
Number of sampled actions with small noises & 2 \\
Number of sampled actions with bigger noises & 2 \\
Reanalyzed ratio & 1.0 \\
\bottomrule
\end{tabular}
}
\end{small}
\end{center}
\captionsetup{labelformat=empty}
    \caption{\textbf{Extended Table. 4:} Hyper-parameters for GridZero}
\end{table}

\end{document}